\pdfoutput=1

\documentclass[journal]{IEEEtran}
\usepackage{times}
\usepackage{epsfig}
\usepackage{graphicx}
\usepackage{amsbsy}
\usepackage{amsmath}
\usepackage{amssymb}
\usepackage{amsthm}
\usepackage{algorithm}
\usepackage{algorithmic}
\usepackage{epstopdf}
\usepackage{multirow}
\usepackage{url}
\usepackage{caption}
\usepackage{subcaption}
\usepackage{color}

\DeclareMathOperator*{\argmin}{arg\,min}
\usepackage{cite}

\ifCLASSINFOpdf
\else
\fi
\begin{document}
\title{Cross-Domain Visual Recognition via Domain Adaptive Dictionary Learning}
\author{Hongyu~Xu,~\IEEEmembership{Student Member,~IEEE},
        Jingjing~Zheng, 
        Azadeh~Alavi,~\IEEEmembership{Member,~IEEE}
        and~Rama~Chellappa,~\IEEEmembership{Fellow,~IEEE}
\thanks{H. Xu, A. Alavi and R. Chellappa are with the Department of Electrical and Computer Engineering, University of Maryland, College Park, MD, 20742, USA.} \thanks{E-mail: \{hyxu, azadeh, rama\}@umiacs.umd.edu} \thanks{J. Zheng is with General Electric Global Research, 1 Research Circle, Niskayuna, NY, 12309, USA.} \thanks{E-mail: {jingjing.zheng}@ge.com}}
\IEEEtitleabstractindextext{%
\begin{abstract}
In real-world visual recognition problems, the assumption that the training data (source domain) and test data (target domain) are sampled from the same distribution is often violated. This is known as the domain adaptation problem. 
In this work, we propose a novel domain-adaptive dictionary learning framework for cross-domain visual recognition. Our method generates a set of intermediate domains. These intermediate domains form a smooth path and bridge the gap between the source and target domains. Specifically, we not only learn a common dictionary to encode the domain-shared features, but also learn a set of domain-specific dictionaries to model the domain shift. The separation of the common and domain-specific dictionaries enables us to learn more compact and reconstructive dictionaries for domain adaptation. These dictionaries are learned by alternating between domain-adaptive sparse coding and dictionary updating steps. Meanwhile, our approach gradually recovers the feature representations of both source and target data along the domain path. By aligning all the recovered domain data, we derive the final domain-adaptive features for cross-domain visual recognition. Extensive experiments on three public datasets demonstrates that our approach outperforms most state-of-the-art methods.
\end{abstract}

}

\maketitle

\IEEEdisplaynontitleabstractindextext

\IEEEpeerreviewmaketitle

\section{Introduction}\label{sec::introduction}
In most traditional visual recognition problems, classifiers are trained using the training data and then applied on the test data, which are assumed to be sampled from the same distribution. 
However, in real-world scenarios, the assumption that the training data and test data are sampled from the same distribution is often violated. These data are generally acquired from multiple source and modalities. For example, in face and object recognition~\cite{Bodla_BZXCCC_WACV2017} applications, training and testing images may be acquired with different cameras or even under different environments and illuminations. Moreover, in video-based action recognition, one action class in different videos may vary significantly due to variation in the camera viewpoints, background or even subject performing the action differently. Besides, such distribution shifts are also observed in human detection~\cite{Vazquez_VLMPG_TPAMI2014,Xu_XLWRBC_2017} and video concept detection~\cite{Yang_YYH_2007,Duan_DTXM_CVPR2009,Duan_DXTL_CVPR2010}. In all these scenarios, the difference in capturing images results in the different data distribution between training and test data. Directly applying the classifiers trained using the training set to test set often yields poor recognition performance. 

Recently, cross-domain visual recognition has gained lots of popularity in the vision community. It involves two types of domain, \textit{source domain} and \textit{target domain}. The source domain usually comprises sufficient amount of labeled training data, using which a classifier could be learned. The target domain contains large amount of unlabeled data, which has different data distribution from the source domain. How to tackle the problem where data from the source domain and target domain have different underlying distribution is also known as the domain adaptation problem. 
Domain adaptation approaches~\cite{Patel_PGC_SPM2014} can be generally categorized into two main types: non-parametric statistical approaches and feature augmentation-based approaches. Nonparametric statistical methods tackle the cross-domain difference by reducing the distribution discrepancy between domains via Maximum Mean Discrepancy (MMD) measurement, while feature augmentation-based approaches focuses more on learning the transferrable feature representations across domains. 

Maximum Mean Discrepancy (MMD) was first introduced in~\cite{Gretton_GBRSS_JMLR2012} to quantize the distribution discrepancy. A fruitful line of work aimed to minimize the MMD between domains via shared feature representation learning or instance weighting~\cite{Pan_PTKY_IJCAI2009,Duan_DTX_TPAMI2012,Long_LWDSY_ICCV13,Long_LWDSY_CVPR2014}. In~\cite{Pan_PTKY_IJCAI2009}, Pan et al. learned transfer components across domains in a Reproducing Kernel Hilbert Space (RKHS) using MMD, such that the kernel mean of the training data becomes close to that of the test data. Duan et al.~\cite{Duan_DTX_TPAMI2012} proposed to learn a kernel function along with a classifier by minimizing empirical loss and
MMD via multiple kernel learning (MKL). Long et al. presented a joint feature matching and instance reweighting method~\cite{Long_LWDSY_ICCV13,Long_LWDSY_CVPR2014}. More specifically, it integrated the minimization of MMD and Frobenius or $\ell_{2,1}$-norm to construct cross-domain features, which are effective even when substantial domain difference is present. However, all the approaches mentioned above may suffer from several disadvantages since MMD is a kernel method. First, the kernel function is not effective for capturing global nonlinearity of the data as it only characterizes the local generalization property. Second, kernel methods scale quadratically to the number of data samples. These open issues have not been well addressed in the previous works.

Feature augmentation-based approaches focuse more on generating the discriminative features that are common or invariant across both domains~\cite{Qiang_QVTC_ECCV2012,Fernando_FHST_ICCV2013,Shi_SS_ICML2012,Gong_GSSG_CVPR2012,Gopalan_GLC_ICCV2011,Shrivastava_SSP_WACV2014,Ni_NQC_CVPR2013,Baktashmotlagh_BHLS_ICCV13,Baktashmotlagh_BHLS_CVPR2014}. Among them, subspace-based methods have been successfully applied to tackle the domain adaptation problem~\cite{Fernando_FHST_ICCV2013,Shi_SS_ICML2012,Gong_GSSG_CVPR2012,Gopalan_GLC_ICCV2011,Baktashmotlagh_BHLS_ICCV13,Baktashmotlagh_BHLS_CVPR2014,Shrivastava_SSP_WACV2014}. Several promising approaches
focused on developing intermediate feature representations along a virtual path connecting the source and target domains. Gopalan et al.~\cite{Gopalan_GLC_ICCV2011} proposed to generate intermediate subspaces by sampling the geodesic path connecting the source and target subspaces on the Grassmann manifold. Gong et al.~\cite{Gong_GSSG_CVPR2012} further improved this by integrating an infinite number of intermediate subspaces to derive a geodesic flow kernel to model the domain shift. Moreover, in~\cite{Shrivastava_SSP_WACV2014}, Shrivastava et al. proposed a parallel transport of the union of the source domain on the Grassmann manifold to overcome the limitation of the assumption of single domain shift between source and target domains. However, the domain subspaces obtained using principle component analysis (PCA) may not well represent the original data and some useful information for adaptation may be lost. In order to overcome the limitation of the PCA subspaces, several recent works~\cite{Ni_NQC_CVPR2013,Shekhar_SPNC_CVPR2013,Qiang_QVTC_ECCV2012,Xu_XZAC_WACV2016,Xu_XZAC_ICPR2016} proposed to use dictionaries to represent the domain data, as non-orthogonal atoms (columns) in the dictionary provide more flexibility to model and adapt the domain data. More specifically, Qiu et al.~\cite{Qiang_QVTC_ECCV2012} learned a parametric dictionary by aligning dictionaries from both domains. In~\cite{Shekhar_SPNC_CVPR2013}, Shekhar et al. jointly learned the projections of data in two domains, and a latent dictionary which can represent both domains in the projected low-dimensional space. Ni et al.~\cite{Ni_NQC_CVPR2013} generated a set of intermediate domains and dictionaries which smoothly adapt the source domain to the target domain.

\begin{figure}
\centering{
\includegraphics[trim = 0in 11.5in 6in 0in, clip,width = 3.3in,height = 0.76in]{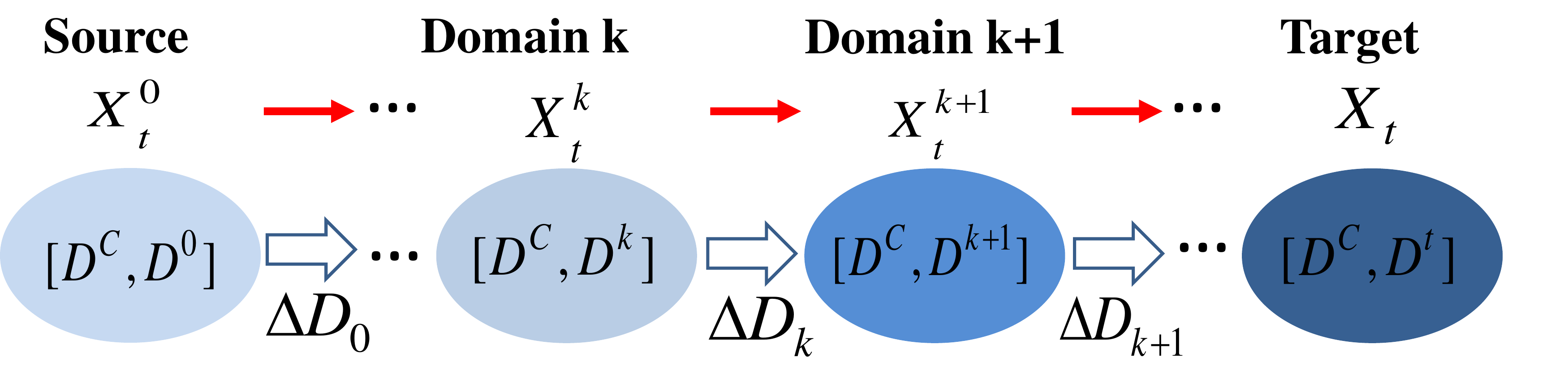}}
\caption{ \small{\bf Our domain-adaptive dictionary learning framework for cross-domain visual recognition.}
The common dictionary $\mathbf{D}^C$ is learned by minimizing the reconstruction error of source and target data. And domain-specific dictionaries $\mathbf{D}^k$ are learned by iteratively adapting the source-specific dictionary $\mathbf{D}^0$ to incrementally reduce the reconstruction error of target data. Meanwhile, at each iterative step when $t =k$, we regularize feature representations of target data in already generated intermediate domains $X_t^i, i \in [0, k-1]$ to have the same sparse codes. This will make all domain changes to be encoded in domain-specific dictionaries $\mathbf{D}^k$.}
\label{fig::motivation_fig}
\end{figure}

In this paper, we tackle the cross-domain visual recognition problem by taking advantage of dictionary learning techniques. This is due to the fact that visual images could be well represented by approximately learned dictionary and corresponding sparse codes. 
Following the approach of creating intermediate feature representations, we propose a novel domain-adaptive dictionary learning approach to generate a set of intermediate domains that bridge the gap between source and target domains. 
Our approach defines two types of dictionaries: a common dictionary and a domain-specific dictionary. The common dictionary shared by all domains is used to extract domain-shared features, whereas the domain-specific dictionary which is incoherent to the common dictionary models the domain shift. The separation of the common dictionary from domain-specific dictionaries will enable us to learn more compact and reconstructive dictionaries for deriving intermediate features.

Here we introduce our domain-adaptive dictionary learning approach (DADL) for cross-domain visual recognition as described in Figure~\ref{fig::motivation_fig}, which aims to generate a set of intermediate dictionaries to gradually reduce the reconstruction error of target data. 
At the beginning, when only source and target data are available, we learn a common dictionary by minimizing the reconstruction error of both source and target data. Then combined with the common dictionary, we learn a set of domain-specific dictionaries by alternating between two steps iteratively: 1) domain adaptive sparse coding: at each iterative step $t = k$, we learn domain-adaptive sparse codes of synthesized feature representations of target data across all available domains. We also regularize feature representations of target data in all the available domains to have the same sparse codes. In this way, we can encode all the domain changes into domain-specific dictionaries. 2) dictionary updating: we update the current domain-specific dictionary to generate the next domain-specific dictionary such that the reconstruction error of target data is further minimized. 
This step not only guarantees that the next domain-specific dictionary will better represent the target data, but also ensures that the intermediate domains gradually adapt to the target domain. Finally, we apply domain-adaptive sparse codes combined with domain dictionaries to construct final domain-adaptive features for cross-domain learning and recognition.

While Ni et al.'s work in~\cite{Ni_NQC_CVPR2013} may be the closest to our work in spirit, our approach differs from~\cite{Ni_NQC_CVPR2013} in the following three aspects: (1) The separation of the common dictionary from domain-specific dictionaries. We aim to learn \textit{both} the common dictionary and domain-specific dictionaries to represent each intermediate domain while~\cite{Ni_NQC_CVPR2013} used only a \textit{single} dictionary to represent each domain. Our approach has two advantages over~\cite{Ni_NQC_CVPR2013}. First, our approach can better represent the domain data because the reconstruction error of domain data obtained using our method is smaller as shown in Figure~\ref{fig::decrease_residue} in Section~\ref{sec::experiments}. Second, the domain-specific dictionaries can better model the domain changes because the domain-shared features are accounted for separately.
(2) The regularization of sparse coding: In each step, we regularize the representation of the target data along the path to have the same sparse codes, which are further used for dictionary updating in the next step. However, the sparse codes used in~\cite{Ni_NQC_CVPR2013} for dictionary updating are only adaptive between \textit{neighboring} domains. Therefore, the sparse representations of target data in~\cite{Ni_NQC_CVPR2013} are \textbf{\textit{not}} domain-adaptive, while the sparse representations in our approach are \textbf{\textit{domain-adaptive}}. Moreover, the intermediate domains generated by our approach are smoother and incorporate the domain change in a better way, which will be verified and discussed in Section~\ref{sec::empirical_analysis}
(3) The construction of final features. We use the domain-adaptive sparse codes across all the  domains multiplied by the dictionaries to represent source and target data, while~\cite{Ni_NQC_CVPR2013} only uses the sparse code decomposed with source and target dictionaries respectively to represent the new features. Therefore, compared to~\cite{Ni_NQC_CVPR2013}, our approach generates more robust and domain-adaptive features.

Recently, adaptive deep neural networks~\cite{Tzeng_THZSD_2014,Long_LCWJ_ICML2015,Ganin_GL_ICML15} have been explored for unsupervised domain adaptation. DDC~\cite{Tzeng_THZSD_2014} minimized Maximum Mean Discrepancy (MMD) via a single linear kernel. DAN~\cite{Long_LCWJ_ICML2015} further extended~\cite{Tzeng_THZSD_2014} by utilizing multiple kernels with multiple layers. In addition, ReverseGrad~\cite{Ganin_GL_ICML15} added a binary classifier to explicitly confuse the source and target domains. It is noted that our approach learns the domain-adaptive feature representations regardless of hand-crafted or deep learning-based features. Moreover, we also evaluate the proposed DADL approach on deep learning-based features to demonstrate the effectiveness of our approach. 

In summary, we make the following contributions:
\begin{itemize}
\item A domain-adaptive dictionary learning framework is proposed for cross-domain visual recognition. We learn a common dictionary to extract features shared by all the domains and a set of domain-specific dictionaries to encode the domain shift. The separation of the common dictionary from domain-specific dictionaries enables us to learn more compact and reconstructive representations for learning.
\item We propose a domain-adaptive sparse coding formulation to incrementally adapt the dictionaries learned from the source domain to reduce the reconstruction error of target data.
\item We recover the feature representations of source and target data in all intermediate domains and extract novel domain-adaptive features by concatenating these intermediate features.
\item We conduct experiments for both face recognition, object classification and active authentication. Both hand-crafted features and deep learning-based features are extensively evaluated. The empirical results demonstrate that the proposed domain-adaptive dictionary learning approach achieves the state-of-the-art performance over most existing cross-domain visual recognition algorithms. 
\end{itemize}

A preliminary version of this work appeared in~\cite{Xu_XZC_BMVC2015}. Theoretical analysis as well as empirical support, extensive experimental evaluation on both hand-crafted features and deep learning-based features and evaluations on active authentication are extensions to~\cite{Xu_XZC_BMVC2015}.

The rest of the paper is organized as follows: Section~\ref{sec::preliminaries} briefly reviews the sparse coding and dictionary learning algorithm. Section~\ref{sec::body} presents the proposed domain-adaptive dictionary learning approach. Section~\ref{sec::optimization} describes the optimization procedure used in the proposed approach. 
Section~\ref{sec::experiments} provides experimental results and analysis on three public datasets. Section~\ref{sec::conclusion} concludes the paper.

\section{Preliminaries}
\label{sec::preliminaries} 
In this section, we give a brief review of sparse coding and dictionary learning as our formulation is based on these topics. 

\subsection{Sparse Coding}
Given a set of $N$ input signals $\mathbf{X} = [x_1,...,x_N] \in \mathbb{R}^{d \times N}$ and a dictionary $\mathbf{D} \in \mathbb{R}^{d \times n}$  of size $n$, the sparse codes $\mathbf{Z}$ of $\mathbf{X}$ are obtained by solving: 
\begin{equation}
\mathbf{Z} = \argmin_{\mathbf{Z}} \| \mathbf{X} - \mathbf{D} \mathbf{Z} \|_F^2 \quad \textrm{s.t.} \quad \forall i, \|z_i\|_0 \leq T,
\end{equation}
where $\| \mathbf{X} \|_F$ denotes the Frobenius norm defined as $\| \mathbf{X} \|_F = \sqrt{\sum_{i,j} |X_{i,j}|^2}$. The sparsity constraint $\|z_i\|_0 \leq T$ requires each signal to have $T$ or fewer atoms in its decomposition with respect to the dictionary $\mathbf{D}$. This can be solved via the orthogonal matching pursuit (OMP) algorithm in~\cite{Tropp_TG_TIT2007}.

\subsection{Dictionary Learning}
Dictionary learning aims to learn an over-complete dictionary $\mathbf{D} \in \mathbb{R}^{d \times n}$ from the set of training samples $\mathbf{X} = [x_1,...,x_N] \in \mathbb{R}^{d \times N}$ along with the corresponding sparse representations $\mathbf{Z} = [z_1,...,z_N] \in \mathbb{R}^{n \times N}$ by solving the following optimization problem over ($\mathbf{D}$, $\mathbf{Z}$) : 
\begin{equation}
\|\mathbf{X} - \mathbf{D} \mathbf{Z}\|_F^2 + \lambda \Phi (\mathbf{Z}) \quad \textrm{s.t.} \quad \|d_i\|_2 = 1 , \forall i \in [1, n]
\end{equation}
where $\mathbf{D} = [d_1,...,d_n] \in \mathbb{R}^{d \times n}$ is the learned dictionary with $n$ atoms (columns), and $\mathbf{Z}= [z_1, ..., z_N]$ are the sparse representations of $\mathbf{X}$ correspondingly. $\lambda$ is a non-negative parameter and $\Phi$ promotes the sparsity constraint. The K-SVD algorithm~\cite{Aharon_AEB_TSP2006} is well known for efficiently learning an over-complete dictionary if $\ell_{0}$-norm is enforced. When the $\ell_{1}$-norm is enforced on $\mathbf{X}$ to enforce sparsity, it could be solved as described in~\cite{Mairal_MBPS_JMLR2010}. In the next section, we will formulate the our domain adaptive dictionary learning framework which can be effectively solved based on the K-SVD algorithm. 

\section{Cross-domain Visual Recognition via Domain Adaptive Dictionary Learning}
\label{sec::body}

\subsection{Problem Formulation}

\begin{figure}[t]
\centering{
\includegraphics[trim = 0in 9in 7in 0in, clip,width = 3.3in,height = 1.6in]{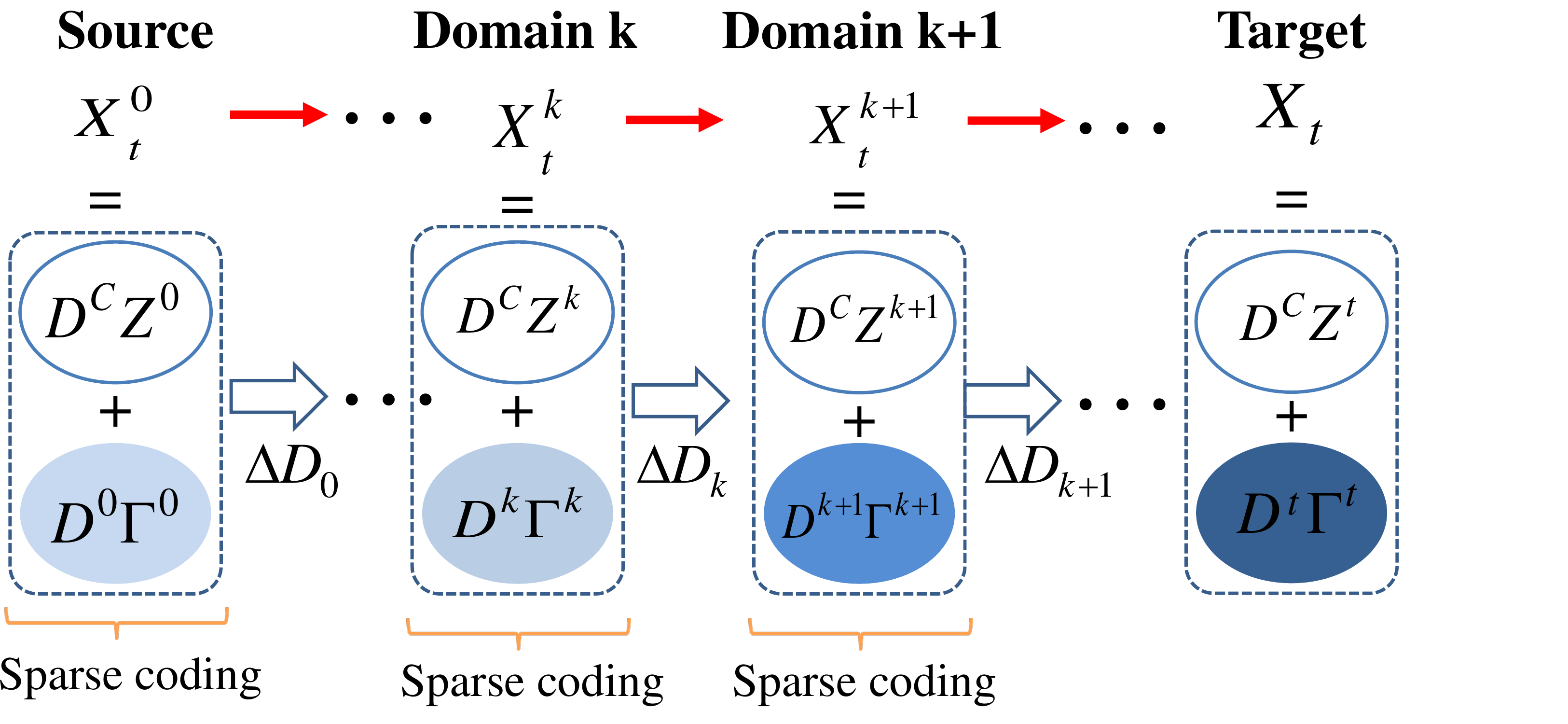}}
\caption{ \small{\bf Details of the learning process of our approach.} The overall learning process consists of three steps: (1) Dictionary learning in source and target domains. At the beginning, we first learn the common dictionary $\boldsymbol{D}^C$, domain-specific dictionaries $\boldsymbol{D}^0$ and $\boldsymbol{D}^t$ for source and target domains. (2) Domain-adaptive sparse coding. At the $k$-th step, we enforce the recovered feature representations of target data in all available domains to have the same sparse codes, while adapting the newest obtained dictionary $\boldsymbol{D}^k$ to better represent the target domain. Then we multiply dictionaries in the $k$-th domain with the corresponding sparse codes to recover feature representations of target data $\boldsymbol{X}_t^k$ in this domain. (3) Dictionary updating. We update $\boldsymbol{D}^k$ to find the next domain-specific dictionary $\boldsymbol{D}^{k+1}$ by further minimizing the reconstruction error in representing the target data. Then we alternate between sparse coding and dictionary updating steps until the stopping criteria is satisfied.}
\label{fig::illustration_fig}
\end{figure}

Let $\mathbf{X}_{s}\in\mathbb{R}^{d\times N_{s}}$ and $\mathbf{X}_{t}\in\mathbb{R}^{d\times N_{t}}$ be the feature representations for source and target domains respectively, where $d$ is the feature dimension, $N_{s}$ and $N_{t}$ are the number of samples in the two domains. Similarly, we denote the feature representations of source and target data in the $k$-th intermediate domain as $\mathbf{X}_s^k \in\mathbb{R}^{d\times N_{s}}$ and $\mathbf{X}_t^k \in\mathbb{R}^{d\times N_{t}}$ respectively.  The common dictionary shared by all the domains is denoted as $\mathbf{D}^C$, whereas source-specific and target-specific dictionaries are denoted as $\mathbf{D}^{0}$, $\mathbf{D}^t$ respectively. Similarly, we use $\mathbf{D}^k, k = 1...N$ to denote the domain-specific dictionaries from the intermediate domains, where $N$ indicates the number of intermediate domains. It is nontrivial to note that, we let all the dictionaries to be of the same size $\in\mathbb{R}^{d\times n}$. Specifically each domain is represented by both the common dictionary and its own domain-specific dictionary. The overall learning process is shown in Figure~\ref{fig::illustration_fig}. 


Our objective is to learn the common dictionary $\mathbf{D}^C$ and a set of domain-specific dictionaries $\mathbf{D}^k$ for generating intermediate domains. Starting from source-specific dictionary $\mathbf{D}^0$ in the source domain, we sequentially learn the intermediate domain-specific dictionaries $\{\mathbf{D}^{k}\}_{k=1}^{N}$ to gradually reduce the reconstruction error of the target data. Figure~\ref{fig::illustration_fig} illustrates the learning process through three steps: (1) Dictionary learning in source and target domain. At the beginning, we first learn the common dictionary $\mathbf{D}^C$ and two domain-specific dictionaries  $\mathbf{D}^0$, $\mathbf{D}^t$ for the source and target domains respectively. (2) Domain-adaptive sparse coding. At the $k$-th step, we learn domain-adaptive sparse codes of target data and recover the feature representations of target data in the $k$-th domain. (3) Dictionary updating. We update the current domain-specific dictionary $\mathbf{D}^k$ to find the next domain-specific dictionary $\mathbf{D}^{k+1}$ by further minimizing the residual error in representing the target data. We alternate between dictionary updating and sparse coding steps until the stopping criteria is satisfied.


\subsection{Dictionary Learning in Source and Target Domains}
At the beginning, we learn the common dictionary $\mathbf{D}^{C}$, source-specific dictionaries $\mathbf{D}^0$ and target-specific dictionary $\mathbf{D}^t$. Given source and target data $\mathbf{X}_{s}$ and $\mathbf{X}_{t}$, we solve for $\mathbf{D}^C$ by minimizing the reconstruction error of both source and target data as follows:
\begin{equation}
\begin{split}
\min_{{\mathbf{D}^{C}}, {\mathbf{Z}^{0}}, {\mathbf{Z}^t}} & ||\mathbf{X}_{s} - \mathbf{D}^{C} \mathbf{Z}^0||_F^{2} + ||\mathbf{X}_{t} - \mathbf{D}^{C} \mathbf{Z}^t||_F^2 \\
& s.t.\hspace{1mm} \forall i,\hspace{2mm} \|z_i^0\|_{0}\leq T, \|z_i^t\|_{0}\leq T
\end{split}
\label{eqn::invariant dictionary}
\end{equation}
where $\mathbf{Z}^0 = [z_1^0...z_{N_s}^0] \in \mathbb{R}^{n\times N_s} , \mathbf{Z}^t = [z_1^t...z_{N_t}^t] \in \mathbb{R}^{n \times N_t}$ are sparse representations of $\mathbf{X}_{s}$ and $\mathbf{X}_{t}$ respectively, $T$ specifies the sparsity that each sample has fewer than $T$ dictionary atoms (columns) in its decomposition.

Given the learned $\mathbf{D}^C$ and corresponding sparse codes $\mathbf{Z}^0$ and $\mathbf{Z}^t$, we learn domain-specific dictionaries $\mathbf{D}^0$ and $\mathbf{D}^t$ by further reducing the reconstruction error of the source and target data. The objective function for learning $\mathbf{D}^0$ and $\mathbf{D}^t$ is given as follows:


\begin{equation}
\begin{split}
&\underset{\mathbf{D}^{0}, \mathbf{\Gamma}^{0}}\min \|\mathbf{X}_{s}-\mathbf{D}^{C}\mathbf{Z}^0 - \mathbf{D}^{0}\mathbf{\Gamma}^0\|_{F}^{2} + \lambda\|\mathbf{D}^{0}{\mathbf{D}^{C}}^T\|_{F}^{2} \\
&\underset{\mathbf{D}^{t}, \mathbf{\Gamma}^{t}}\min \|\mathbf{X}_{t}-\mathbf{D}^{C}\mathbf{Z}^t - \mathbf{D}^{t}\mathbf{\Gamma}^t\|_{F}^{2} + \lambda\|\mathbf{D}^{t}{\mathbf{D}^{C}}^T\|_{F}^{2} \\
& s.t.\forall i,\|z_i^0\|_{0} + \|\alpha_i^0\|_{0}\leq T, \hspace{3mm} \|z_i^t\|_{0}+ \|\alpha_i^t\|_{0}\leq T\\
\end{split}
\label{eqn::source_and_target_dictionries}
\end{equation}

where $\mathbf{\Gamma}^0 = [\alpha_1^0...\alpha_{N_s}^0] \in \mathbb{R}^{n\times N_s}$ and $\mathbf{\Gamma}^t = [\alpha_1^t...\alpha_{N_t}^t] \in \mathbb{R}^{n \times N_t}$ are sparse representations of $\mathbf{X}_{s}$ and $\mathbf{X}_{t}$ with respect to $\mathbf{D}^0$ and $\mathbf{D}^t$, and $\lambda$ is the regularization parameter.
The first term in both functions in~\eqref{eqn::source_and_target_dictionries} is the reconstruction error of domain data using both the common dictionary and corresponding domain-specific dictionary. The second term is the inner product of the atoms from different dictionaries, which encourages $\mathbf{D}^C$ to be incoherent to the domain-specific dictionaries. This incoherence term minimizes the correlation between $\mathbf{D}^{C}$ and $\{\mathbf{D}^{0}, \mathbf{D}^{t}\}$, thus it enables our approach to exploit domain-shared features and domain changes separately. We describe the optimization of the objective functions in~\eqref{eqn::source_and_target_dictionries} in Section~\ref{sec::optimization}.

\subsection{Domain-adaptive Sparse Coding}

At the $k$-th step, assume we have already generated ($k$-1) intermediate domains and domain-specific dictionaries denoted as $\{\mathbf{X}_{t}^{i}\}_{i=1}^{k-1}$ and $\{\mathbf{D}^{i}\}_{i=1}^{k-1}$ respectively. Now given a newly obtained domain-specific dictionary $\mathbf{D}^{k}$ for the $k$-th domain, we want to obtain
sparse representations of target data $\mathbf{X}_t$ in the $k$-th domain. In order to achieve this goal, we not only reconstruct $\mathbf{X}_t$ using dictionaries from the $k$-th domain,
but also reconstruct the recovered target data $\mathbf{X}_t^i$ in each intermediate domain using dictionaries from that domain.
Moreover, we regularize the sparse representation of $\mathbf{X}_s$,  $\mathbf{X}_t$ and $\mathbf{X}_t^i$ to be the same.
This regularization step ensures that the sparse representations of target data across all available domains are the same (i.e. \textbf{\textit{domain-adaptive}}).
We solve for domain-adaptive sparse codes across all the available domains as follows:
\begin{equation}
\small
\begin{split}
\mathbf{Z}^{k},\mathbf{\Gamma}^{k} = & \underset{\mathbf{Z},\mathbf{\Gamma}}\argmin \|\mathbf{X}_{t}-\mathbf{D}^C\mathbf{Z}-\mathbf{D}^{k}\mathbf{\Gamma}\|_{F}^{2}+ \\
& \sum_{i = 0}^{k-1} \|\mathbf{X}_t^{i}-\mathbf{D}^C\mathbf{Z}-\mathbf{D}^i\mathbf{\Gamma}\|_{F}^{2} +  \|\mathbf{X}_{t} - \mathbf{D}^C\mathbf{Z}-\mathbf{D}^t\mathbf{\Gamma}\|_{F}^{2} \\
& \hspace{3mm} s.t. \hspace{2mm} \forall i, \|z_i\|_{0}+ \|\alpha_i\|_{0}\leq T
\label{eqn::optimal sparse code}
\end{split}
\end{equation}
where $\mathbf{Z}^{k} = [z_1^{k}...z_{N_t}^{k}], \mathbf{\Gamma}^{k} = [\alpha_1^{k}...\alpha_{N_t}^{k}]$ are the solved sparse representations of target data in the $k$-th domain, $\mathbf{X}_t^{i} = \mathbf{D}^C\mathbf{Z}^{i} + \mathbf{D}^{i} \mathbf{\Gamma}^{i}$ are the recovered feature representations of target data in the $i$-th domain obtained in previous iteration steps. The objective function in~\eqref{eqn::optimal sparse code} has three terms:
\begin{enumerate}
\item  The first term is the reconstruction error of target data when encoded using dictionaries from the $k$-th domain. This term is called \textbf{\textit{domain shifting}} term, because it adapts dictionaries in the $k$-th domain to better represent the target data.

\item The second term in~\eqref{eqn::optimal sparse code} sums the reconstruction errors of recovered feature representations of target data in all the intermediate domains. 

\item The last term is the reconstruction error of target data in the target domain.
\end{enumerate}

These two terms above are called \textbf{\textit{domain adaptive}} terms. This is because we regularize both $\mathbf{X}_t$ and $\mathbf{X}_t^i$ to have the same sparse codes. It means that feature representations of recovered target data in different domains will have the same sparse codes when encoded using dictionaries from each domain. This regularization will guarantee that sparse codes are domain-adaptive, such that the domain changes are encoded only in domain-specific dictionaries.

Then we recover the feature representations of target data in the $k$-th domain $\mathbf{X}_t^{k}$ as follows: $\mathbf{X}_t^{k} = \mathbf{D}^C\mathbf{Z}^{k} + \mathbf{D}^{k} \mathbf{\Gamma}^{k}$.

\subsection{Domain-specific Dictionary Updating}

After sparse coding at the $k$-th step,  we will update $\mathbf{D}^k$ to find the next domain-specific dictionary $\mathbf{D}^{k+1}$ by further reducing the reconstruction error of target data in the $k$-th domain. Let $\mathbf{J}^k$ denote the target reconstruction residue in the $k$-th domain, which is computed as follows:
\begin{equation}
\mathbf{J}^{k} = \mathbf{X}_{t} - \mathbf{D}^{C}\mathbf{Z}^{k} - \mathbf{D}^{k}\mathbf{\Gamma}^{k} \\
\label{eqn::residue target}
\end{equation}
where $\mathbf{Z}^{k}$ and $\mathbf{\Gamma}^{k}$ are the sparse codes obtained for reconstructing $\mathbf{X}_t$ in the $k$-th step. We further reduce the target reconstruction residue $\mathbf{J}^k$ by adjusting $\mathbf{D}^k$ by $\Delta\mathbf{D}^k \in \mathbb{R}^{d \times n}$, which is solved from the following problem:

\begin{equation}
\underset{\Delta\mathbf{D}^{k}}\min \|\mathbf{J}^{k}-\Delta\mathbf{D}^{k}\mathbf{\Gamma}^{k}\|_{F}^{2} + \eta\|\Delta\mathbf{D}^{k}\|_{F}^{2} \\
\label{eqn::objective_deltaD}
\end{equation}
The objective function in~\eqref{eqn::objective_deltaD} has two terms. The first term ensures that the adjustment $\Delta \mathbf{D}^k$ will further reduce the target reconstruction residue $\mathbf{J}^k$. While the second term penalizes the abrupt changes between two adjacent domain-specific dictionaries so that the intermediate domains smoothly adapt to the target domain. The parameter $\eta$ controls the balance between these two terms. Since the problem in~\eqref{eqn::objective_deltaD} is a ridge regression problem, we solve for $\Delta\mathbf{D}^k$ by setting the first derivative to be zeros as in~\cite{Ni_NQC_CVPR2013} and obtain:
\begin{equation}
\Delta\mathbf{D}^{k} = \mathbf{J}^{k}{\mathbf{\Gamma}^{k}}^{T}(\eta\mathbf{I}+\mathbf{\Gamma}^{k}{\mathbf{\Gamma}^{k}}^{T})^{-1}
\label{eq::dictionary_update_1}
\end{equation}
where $\mathbf{I}\in \mathbb{R}^{n\times n}$ is the identity matrix. The next domain-specific dictionary $\mathbf{D}^{k+1}$ is obtained as: $\mathbf{D}^{k+1} = \mathbf{D}^{k} + \Delta \mathbf{D}^{k}$. In addition, we normalize each column in $\mathbf{D}^{k+1}$ to be a unit vector. 

\newtheorem{proposition}{Proposition}
\begin{proposition}
The residue $\mathbf{J}^{k}$ in~\eqref{eqn::residue target} is non-increasing with respect to $\mathbf{D}^C$, $\mathbf{D}^{k}$, $\mathbf{\Delta D}^{k}$ and corresponding sparse codes $\mathbf{Z}^{k},\mathbf{\Gamma}^{k}$, \textit{i.e.} $\|\mathbf{J}^{k} - \Delta\mathbf{D}^{k}\mathbf{\Gamma}^{k}\|_{F}^{2} \leq \|\mathbf{J}^{k}\|_{F}^{2}$. \\
\label{prop::1}
\end{proposition}

The non-increasing property of the residue $\mathbf{J}^k$ ensures that the source-specific dictionary $\mathbf{D}^0$ gradually adapts to the target-specific dictionary $\mathbf{D}^t$ through a set of intermediate domain-specific dictionaries $\mathbf{D}^k$. The proof is given in the appendix.

After the domain-specific dictionary update, we increase $k$ by $1$, and alternate between the sparse coding step in section $3.2$ and the dictionary updating step in section $3.3$ until the stopping criteria is reached. We summarize our approach in Algorithm $1$.

\begin{algorithm}[!tb]
\caption{The proposed domain adaptive dictionary learning framework}
\label{alg:overall}
\begin{algorithmic}[1]
\STATE \textbf{Input:} source data $\mathbf{X}_s$ , target data $\mathbf{X}_t$, sparsity level $T$, parameter $\lambda$, $\eta$, stopping threshold $\delta$
\STATE \textbf{Output:} $\mathbf{D}^C$, $\mathbf{D}^{0}$ and $\mathbf{D}^{t}$
\STATE Compute $\mathbf{D}^C$ using ~\eqref{eqn::invariant dictionary}
\STATE Compute $\mathbf{D}^{0}$, $\mathbf{D}^{t}$ by solving the objective function in~\eqref{eqn::source_and_target_dictionries}.
\STATE $k=0$
\WHILE{stopping criteria is not reached}
\STATE compute domain-adaptive sparse codes $\mathbf{Z}^{k}$, $\mathbf{\Gamma}^{k}$ using equation~\eqref{eqn::optimal sparse code}
\STATE compute the reconstruction error $\mathbf{J}^{k}$ using equation~\eqref{eqn::residue target}.
\STATE compute the adjustment $\Delta\mathbf{D}^{k}$ using equation~\eqref{eq::dictionary_update_1}
\STATE $\mathbf{D}^{k+1} \leftarrow \mathbf{D}^{k} + \Delta \mathbf{D}^{k}$
\STATE normalize $\mathbf{D}^{k+1}$ to have unit atoms.
\STATE $\mathbf{X}_{t}^{k+1} \leftarrow \mathbf{D}^C\mathbf{Z}^{k} + \mathbf{D}^{k} \mathbf{\Gamma}^{k}$
\STATE $ k \leftarrow k+1 $
 \STATE Check the stopping criteria $ \|\Delta\mathbf{D}^{k}\|_{F} \leq \delta$ 
\ENDWHILE
\STATE \textbf{Final Output:} $\mathbf{D}^C$, $\mathbf{D}^{k}, k\in[0,N]$ and $\mathbf{D}_{t}$.
\end{algorithmic}
\end{algorithm}

\subsection{Derivation of New Features for Domain Data}

Until now we have obtained the common dictionary $\mathbf{D}^{C}$, domain-specific dictionaries $\mathbf{D}^{k}, k\in[0,N]$.
The transition path made up of $\mathbf{D}^c$ and the set of domain-specific dictionaries $\mathbf{D}^k$ models the domain shift.
We will make use of it to derive new domain-adaptive representations for source and target data.

Since the recovered feature representations of target data $\mathbf{X}_t^k, k\in[0,N]$ in all intermediate domains are already available, we first recover feature representations of source data $\mathbf{X}_s^k, k\in [0,N]$ in each intermediate domain. We iteratively recover $\mathbf{X}_s^k$ in a similar way as $\mathbf{X}_t^k$. The only difference is that all the dictionaries are already learned and fixed during the learning of $\mathbf{X}_s^k$. Specifically, at the $k$-th iterative step, we obtain the sparse representations of source data that are adaptive across all domains by solving the following problem:
\begin{equation}
\small
\begin{split}
& \mathbf{Z}_{s}^{k},\mathbf{\Gamma}_{s}^k = \argmin_{\mathbf{Z},\mathbf{\Gamma}}  \|\mathbf{X}_{s} - \mathbf{D}^C\mathbf{Z}-\mathbf{D}^t\mathbf{\Gamma}\|_{F}^{2} \\
& + \sum_{i = 1}^{k-1} \|\mathbf{X}_{s}^i-\mathbf{D}^C\mathbf{Z}-\mathbf{D}^i\mathbf{\Gamma}\|_{F}^{2} \hspace{2mm} s.t. \forall i, \hspace{2mm}  \|z_{i}\|_{0} + \|\alpha_{i}\|_{0}\leq T
\end{split}
\label{eqn::final source sparse codes}
\vspace{-1mm}
\end{equation}
where $\mathbf{Z}_s^k = [z_{s_1}^k...z_{s_{N_s}}^k], \mathbf{\Gamma}_s^k = [\alpha_{s_1}^k...\alpha_{s_{N_s}}^k]$ are sparse representations of source data in the $k$-th domain, $\mathbf{X}_s^i = \mathbf{D}^C\mathbf{Z}_s^i + \mathbf{D}^i\mathbf{\Gamma}_s^i$ are recovered feature representations of source data in the $i$-th domain obtained in previous iteration steps. The objective function in~\eqref{eqn::final source sparse codes} consists of two terms. The first term is the reconstruction error of source data using dictionaries from the target domain while the second term is the sum of reconstruction error of recovered feature representations of source data in all intermediate domains. Similarly, we enforce both $\mathbf{X}_s^0$ and $\mathbf{X}_s^i$ to have the same sparse codes. After sparse coding in the $k$-th step, we recover the feature representations of source data in the $k$-th domain as follows: $\mathbf{X}_s^k = \mathbf{D}^C\mathbf{Z}_s^k + \mathbf{D}^k\mathbf{\Gamma}_s^k$.

We use the sparse codes obtained in the last iterative step to derive the new feature representations for the source and target data. The new augmented feature representation of source and target data are $\mathbf{\tilde{X}}_s = [\mathbf{\tilde{X}}_s^0, ..., \mathbf{\tilde{X}}_s^N]$ and $\mathbf{\tilde{X}}_t = [\mathbf{\tilde{X}}_t^0, ..., \mathbf{\tilde{X}}_t^N]$ respectively, where $\mathbf{\tilde{X}}_s^i = \mathbf{D}^C \mathbf{Z}_s^{N}+\mathbf{D}^{i} \mathbf{\Gamma}_s^{N}$ and $\mathbf{\tilde{X}}_t^i = \mathbf{D}^C \mathbf{Z}_t^{N} + \mathbf{D}^{i} \mathbf{\Gamma}_t^{N}$  and
$\mathbf{Z}_s^N$, $\mathbf{Z}_t^N$, $\mathbf{\Gamma}_s^N$, $\mathbf{\Gamma}_t^N$ are the sparse codes obtained in the last iterative step where $k = N$. The final stage of recognition across all the domains is performed using an SVM classifier trained on new feature vectors after dimension reduction via the Principal Component Analysis (PCA).

\section{Optimization}
\label{sec::optimization}
In this section, we provide the details of optimizing the objective functions in~\eqref{eqn::source_and_target_dictionries} and~\eqref{eqn::optimal sparse code}.

\subsection{Source and Target Domain-specific Dictionaries Learning}

The objective functions that we learn to construct the source and target domain-specific dictionaries in~\eqref{eqn::source_and_target_dictionries} could be divided into two subproblems: (a) computing sparse codes with fixed $\mathbf{D}^C$, $\mathbf{D}^0$ and $\mathbf{D}^t$; (b) updating $\mathbf{D}^0$ and $\mathbf{D}^t$ with fixed sparse codes and $\mathbf{D}^C$. Since the two objective functions in~\eqref{eqn::source_and_target_dictionries} are of the same form, we only describe the optimization of $\mathbf{D}^{0}$ as shown below:
\begin{equation}
\begin{split}
& \underset{\mathbf{D}^{0}, \mathbf{Z}^{0}, \mathbf{\Gamma}^{0}}\min \|\mathbf{X}_{s}-\mathbf{D}^{C}\mathbf{Z}^{0} - \mathbf{D}^{0}\mathbf{\Gamma}^0\|_{F}^{2} + \lambda\|\mathbf{D}^{0}{\mathbf{D}^{C}}^T\|_{F}^{2} \\
& \hspace{+9mm} s.t. \hspace{+3mm}\forall i,\|z_i^0\|_{0}+ \|\alpha_i^0\|_{0}\leq T
\end{split}
\label{eqn::source_specified_dictionry}
\end{equation}
(a) Given fixed $\mathbf{D}^C$, $\mathbf{D}^0$,~\eqref{eqn::source_specified_dictionry} is reduced to:
\begin{equation}
\begin{split}
&\underset{\mathbf{Z}^{0}, \mathbf{\Gamma}^{0}}\min \|\mathbf{X}_{s}- \left[\mathbf{D}^C \hspace{1mm} \mathbf{D}^0 \right] \left[
                                            \begin{array}{c}
                                                \mathbf{Z}^{0}\\
                                                \mathbf{\Gamma}^{0} \\
                                                \end{array}
                                            \right] \|_{F}^{2} \\
& \hspace{+3mm} s.t. \hspace{+3mm}\forall i,\|z_i^0\|_{0}+ \|\alpha_i^0\|_{0}\leq T\\
\end{split}
\label{eqn::compute_sparse_code}
\end{equation}
The minimization of \eqref{eqn::compute_sparse_code} is a LASSO problem and we compute $\mathbf{Z}^0, \mathbf{\Gamma}^{0}$ using the method proposed in \cite{Mairal_MBPS_JMLR2010}. \\
(b) Given fixed sparse coefficients $\mathbf{Z}^{0}, \mathbf{\Gamma}^{0}$,~\eqref{eqn::source_specified_dictionry} is reduced to:
\begin{equation}
\begin{split}
\underset{\mathbf{D}^{0}}\min \|\mathbf{J}_{s}- \mathbf{D}^{0}\mathbf{\Gamma}^0\|_{F}^{2} + \lambda\|\mathbf{D}^{0}{\mathbf{D}^{C}}^T\|_{F}^{2} \\
\end{split}
\label{eqn::update source specified dictionary}
\end{equation}
where $\mathbf{J}_{s} = \mathbf{X}_s - \mathbf{D}^C \mathbf{Z}^{0}$ is a fixed matrix. Motivated by \cite{Kong_KW_ECCV12} and~\cite{Zheng_ZJ_ICCV2013}, we update $\mathbf{D}^0 = [d_{1}^0,...,d_{n}^0]$ atom by atom, i.e. updating the $j$-th atom $d_{j}^0$ by fixing other atoms in $\mathbf{D}^0$. Specifically, let $\mathbf{\hat{J}_s} = \mathbf{J}_s - \sum_{j \neq k}{d_{j}^0\alpha_{(j)}^0 }$ where $\alpha_{(j)}^0$ corresponds to the $j$-th row of $\mathbf{\Gamma}^0$, then we solve the following problem for updating $d_{j}^0$ in $\mathbf{D}^0$: 
\begin{equation}
d_{j}^0 = \argmin_{d_{j}^0} f(d_{j}^0) = \| \mathbf{\hat{J}}_s - d_{j}^0\alpha_{(j)}^0\|_{F}^{2} + \lambda\|{d_{k}^0}^{T}\mathbf{D}^{C}\|_{F}^{2} 
\label{eqn::update_d_j_0}
\end{equation}

Let the first-order derivative of~\eqref{eqn::update_d_j_0} with respect to $d_{j}^0$ equal to zero, \textit{i.e.} $\frac{\partial f(d_{j}^0)}{\partial d_{j}^0} = 0$, then the closed form solution of $d_{j}^0$ is obtained as:

\begin{equation}
\begin{split}
d_{j}^0 = (\|\alpha_{(j)}^0\|_2^2 \hspace{1mm} \mathbf{I} + \lambda \mathbf{D}^C {\mathbf{D}^C}^T)^{-1}\mathbf{\hat{J}_s}{\alpha_{(j)}^0}^T
\end{split}
\label{eqn::update_atom}
\end{equation}

Also note that as an atom of a dictionary, it should be normalized to unit vector, i.e. $ \hat{{d}_{k}^0}= d_{k}^0 / \|d_{k}^0\|_2$. Along with this, the corresponding coefficient should be multiplied by $\|d_{k}^0\|_2$, \textit{i.e.} $\hat{{\alpha}_{(j)}^0} = \|d_{j}^0\|_2 \hspace{1mm}\alpha_{(j)}^0$.

We alternate between sparse coding and dictionary updating steps until the objective function in~\eqref{eqn::source_specified_dictionry} converges, yielding the source domain-specific dictionary $\mathbf{D}^{0}$ and corresponding sparse coefficients $\mathbf{Z}^{0}, \mathbf{\Gamma}^{0}$.

\subsection{Computing Domain-adaptive Sparse Codes}

In~\eqref{eqn::optimal sparse code}, given fixed $\mathbf{D}^C$ and domain-specific dictionaries $\mathbf{D}^k$, $k \in[0,K]$, the objective function~\eqref{eqn::optimal sparse code} could be rewritten as follows:
\begin{equation}
\mathbf{Z}^{k},\mathbf{\Gamma}^{k} = \underset{\mathbf{Z},\mathbf{\Gamma}}\argmin \|\mathbf{\tilde{X}}-\mathbf{\tilde{D}}\left[ \hspace{1mm}\mathbf{Z} \hspace{2mm} \mathbf{\Gamma} \hspace{1mm} \right]^{T} \|_{F}^{2}
\label{eqn::sparse code optimization}
\end{equation}
where $\mathbf{\tilde{X}} = \left[ \mathbf{X}_t^{T}, \mathbf{X}_t^{T }, {\mathbf{X}_t^{0}}^{T}, ..., {\mathbf{X}_{t}^{k-1}}^{T} \right]^{T}$ and $ \mathbf{\tilde{D}} = {\left[
                            \begin{array}{cc}
                            {\mathbf{D}^t}^T,  {\mathbf{D}^k}^T,  {\mathbf{D}^0}^T, ..., {\mathbf{D}^{k-1}}^T  \\
                            {\mathbf{D}^c}^{T}, {\mathbf{D}^c}^{T}, {\mathbf{D}^c}^{T}, ..., {\mathbf{D}^c}^{T} \\
                            \end{array}
                       \right]}^{T}$.
We can solve~\eqref{eqn::sparse code optimization} as a LASSO problem to compute the sparse codes as in~\cite{Mairal_MBPS_JMLR2010}.

\section{Experiments}
\label{sec::experiments}
In this section, we evaluate our method for three different cross-domain visual recognition tasks.
We use the CMU-PIE dataset introduced in~\cite{Sim_SBB_TPAMI2003}  for face recognition, Office dateset in~\cite{Saenko_SKFD_ECCV2010} for object classification and UMD Active Authentication (UMDAA-$01$) dataset introduced in~\cite{Zhang_ZPFC_WACV2015} for active authentication. We also compare the performance of the proposed approach with several state-of-the-art approaches for cross-domain visual recognition. 

\subsection{Cross-domain Face Recognition}

The CMU-PIE dataset~\cite{Sim_SBB_TPAMI2003} is a controlled face dataset of $68$ subjects with a total of $41,368$ images. Each subject has $13$ images under $9$ different poses, $21$ different illuminations and $4$ different expressions. We perform two series of experiments on cross-domain face recognition tasks. First we evaluate our approach on the task of face recognition across blur and illuminations. Then we carry out experiment of face recognition across pose variation.


\subsubsection{Face Recognition Across Blur and Illuminations}

\begin{figure}[htp!]
\centering{
\includegraphics[scale = 0.24]{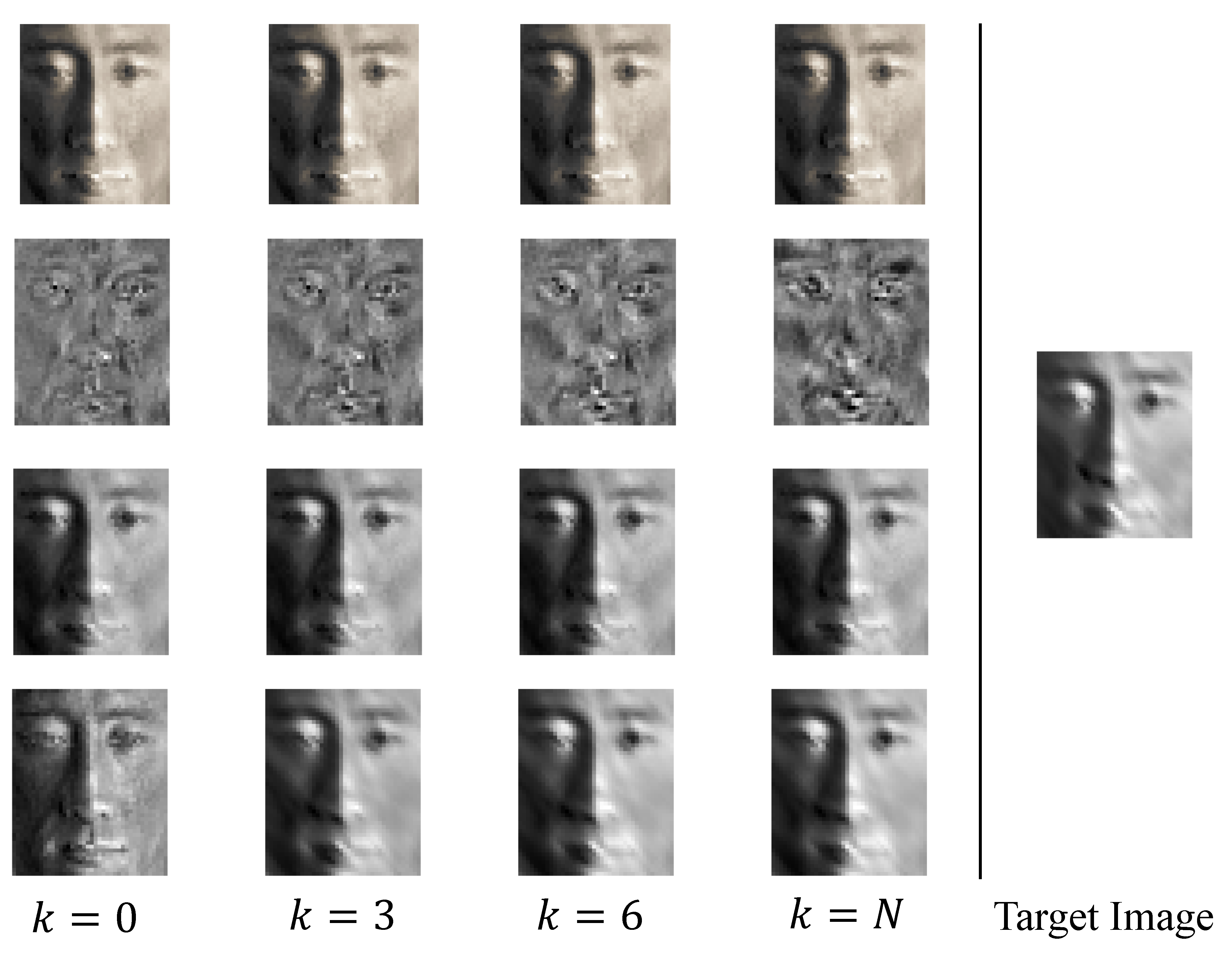}}
\caption{ \small{\bf Synthesized face images of a target face along the intermediate domains.} The image in the last column denote the original target face, while the images in the first four columns are synthesized face images of the target face along four different intermediate domains. The images in row $1$ and row $2$ are components of our synthesized face images corresponding to the common dictionary and the domain-specific dictionaries respectively. The face images in row $3$ and row $4$ are synthesized face images generated by our approach and~\cite{Ni_NQC_CVPR2013} respectively.}
\label{fig::blur_face}
\end{figure}

In this set of experiment, we choose $34$ subjects under first $11$ illumination conditions to compose the source domain. The target domain was formed by the remaining images with the other $10$ illumination conditions. The images in the source domain were labeled, but not those in the target domain. We followed the protocol discussed in~\cite{Ni_NQC_CVPR2013} to synthesize the domain shift by applying two different types of blur kernels to the target data: 1) Gaussian blur kernel with different standard deviations from $2$ to $8$, and 2) motion blur kernel with different lengths from $3$ to $19$ along $\Theta =135^o$. In summary, the domain shift consist of two components. The first is a change in illumination direction, 11 illumination directions in the source domain
and other 10 different illuminations directions in the target domain. The second component is due to blur. Therefore, we evaluate the proposed approach on cross-domain face recognition across blur and illuminations. 


\textbf{Methods for Comparison}: Since our method is based on dictionary learning, we compare our method with several baseline and state-of-the-art cross-domain recognition algorithms listed below: 
\begin{itemize}
\item K-SVD~\cite{Aharon_AEB_TSP2006}. We learn dictionaries from the source domain using the K-SVD algorithm in~\cite{Aharon_AEB_TSP2006}. Then we directly decomposes the target domain data with dictionaries and use the nearest neighbor classifier on the resulting sparse codes for testing. 
\item GFK~\cite{Gong_GSSG_CVPR2012} introduced the geodesic flow kernel on Grassmann manifold to calculate the distance between the source and the target samples.
\item TJM~\cite{Long_LWDSY_CVPR2014} jointly performed feature matching and instance re-weighting across domains.
\item SIDL~\cite{Ni_NQC_CVPR2013} created a set of intermediate dictionaries to model domain shift.
\item CORAL~\cite{Sun_SFS_AAAI2016} minimized the domain shift by aligning the second-order statistics of source and target distribution.
\end{itemize}
In addition, we also compare with other two methods proposed in~\cite{Ahonen_AROH_ICPR2008, Biswas_BAC_TPAMI2009}. ~\cite{Ahonen_AROH_ICPR2008} introduced a blur insensitive descriptor which was called the Local Phase Quantization (LPQ) while~\cite{Biswas_BAC_TPAMI2009} estimated an illumination robust albed map (Albedo) for matching.

Tables~\ref{tab::gaussian} and~\ref{tab::motion} show the classification accuracies of different methods for face recognition across Gaussian blur and motion blur respectively. From the results, the proposed method achieves the best performance for the tasks of face recognition across blur and illuminations. This shows the effectiveness of our method in bridging the domain shift by iterative domain dictionary learning. It is also interesting to note that the baseline K-SVD poorly handles the domain shift between training and testing sets performs poorly, while other cross-domain cross-domain recognition methods improve upon it. In addition, since both illumination and blur variations contribute to the domain shift, LPQ~\cite{Ahonen_AROH_ICPR2008} which is only blur robust and albedo~\cite{Biswas_BAC_TPAMI2009} which is only illumination insensitive are not able to handle all the domain changes. Moreover, our method outperforms~\cite{Ni_NQC_CVPR2013}, the method most similar to ours, which also learned a set of dictionaries to model the domain shift. This is because~\cite{Ni_NQC_CVPR2013} only regularizes two adjacent domains to have the same sparse codes and the learned dictionaries do not fully capture the domain changes. However, our method encodes the domain changes into domain-specific dictionaries well by encouraging feature representation of different domain data to have domain adaptive sparse codes.

\begin{table}[t]
\caption{{\bf Recognition accuracies across different Gaussian blur kernels on the CMU-PIE dataset \cite{Sim_SBB_TPAMI2003}.} Each column corresponds to Gaussian kernels with the standard deviation $\sigma = 2,3,4,5,6,7,8$.}
\label{tab::gaussian}
\centering{
\resizebox{0.47\textwidth}{!}{
\begin{tabular}{|c|c|c|c|c|c|c|c|}
\hline
$\sigma$ & $2$ & $3$ & $4$ & $5$ & $6$ & $7$ & $8$  \\
\hline
Ours      & \textbf{88.9} & \textbf{85.3} & \textbf{84.8} & \textbf{82.7} & \textbf{81.2} & \textbf{80.5} & \textbf{80.7} \\
\hline
CORAL~\cite{Sun_SFS_AAAI2016}          & 83.2 & 78.9 & 76.5 & 74.8 & 74.0 & 72.1 & 62.4 \\  
\hline
TJM \cite{Long_LWDSY_CVPR2014}               & 67.4 & 65.6 & 65.3 & 64.4 & 63.8 & 63.8 & 63.5 \\
\hline
GFK \cite{Gong_GSSG_CVPR2012}            & 81.1 & 78.5 & 77.6 & 75.9 & 74.0 & 72.1 & 70.4  \\
\hline
SIDL \cite{Ni_NQC_CVPR2013}             & 84.0 & 80.3 & 78.9 & 78.2 & 77.9 & 76.5 & 74.8 \\
\hline
LPQ \cite{Ahonen_AROH_ICPR2008}          & 69.1 & 66.5 & 64.4 & 61.6 & 58.3 & 55.3 & 53.2  \\
\hline
Albedo \cite{Biswas_BAC_TPAMI2009}       & 72.4 & 50.9 & 36.8 & 24.8 & 19.6 & 17.3 & 15.7 \\
\hline
K-SVD \cite{Aharon_AEB_TSP2006}        & 49.1 & 41.2 & 36.8 & 34.6 & 32.7 & 29.2 & 28.0  \\
\hline
\end{tabular}}
}
\end{table}

\begin{table}[t]
\caption{{\bf Recognition accuracies across different motion blur kernels on the CMU-PIE dataset \cite{Sim_SBB_TPAMI2003}.} Each column corresponds to motion blur with the length $L = 3,7,9,11,13,15,17$.}
\label{tab::motion}
\centering{
\resizebox{0.47\textwidth}{!}{
\begin{tabular}{| c | c | c | c | c | c | c | c |}
\hline
$L$ & $3$ & $7$ & $9$ & $11$ & $13$ & $15$ & $17$  \\
\hline
Ours      & \textbf{97.9} & \textbf{89.7} & \textbf{88.2} & \textbf{82.5} & \textbf{77.4} & \textbf{75.0} & \textbf{70.8} \\
\hline
CORAL~\cite{Sun_SFS_AAAI2016}        & 87.9 & 77.9 & 70.9 & 68.8 & 65.6 & 61.2 & 59.4 \\
\hline
TJM \cite{Long_LWDSY_CVPR2014}               & 71.8 & 69.4 & 66.2 & 64.1 & 60.0 & 57.1 & 54.1 \\
\hline
GFK \cite{Gong_GSSG_CVPR2012}            & 91.3 & 84.9 & 82.4 & 77.6 & 70.7 & 66.9 & 59.8  \\
\hline
SIDL \cite{Ni_NQC_CVPR2013}             & 95.6 & 86.5 & 85.9 & 81.2 & 75.7 & 72.3 & 63.2 \\
\hline
LPQ \cite{Ahonen_AROH_ICPR2008}          & 81.8 & 77.4 & 73.8 & 62.6 & 54.5 & 47.1 & 43.4   \\
\hline
Albedo \cite{Biswas_BAC_TPAMI2009}       & 82.3 & 70.7 & 60.9 & 45.9 & 35.1 & 26.4 & 18.9 \\
\hline
K-SVD \cite{Aharon_AEB_TSP2006}        & 85.0 & 56.5 & 42.6 & 30.3 & 25.9 & 19.8 & 17.3 \\
\hline
\end{tabular}}
}
\end{table}

\begin{figure*}[htp!]
\centering{
\includegraphics[scale = 0.38]{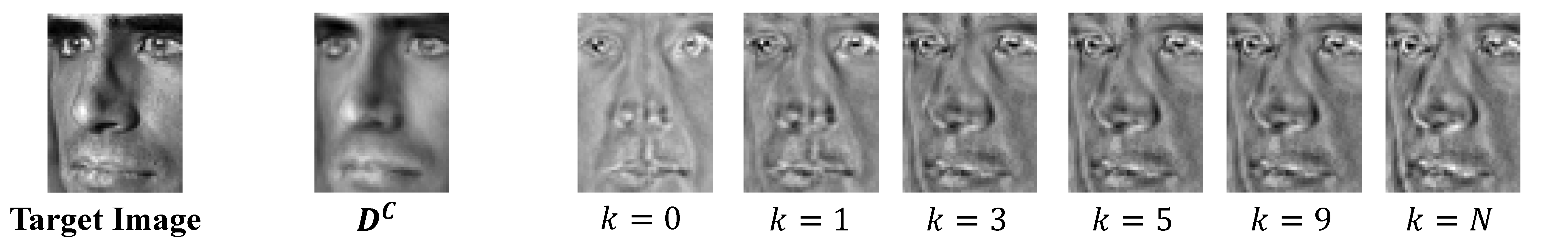}}
\caption{ {\bf Recovered face images of a target face image along the intermediate domains.} The first image is the original target face image, the second image is the component of the recovered face image corresponding to the common dictionary. The remaining six images are the components of the recovered face images corresponding to domain-specific dictionaries.}
\label{fig::pose_face}
\end{figure*}

\begin{table}[t]
\caption{{\bf Recognition accuracies across pose variation on the CMU-PIE dataset \cite{Sim_SBB_TPAMI2003}.}}
\label{tab::pose}
\centering{
\resizebox{0.47\textwidth}{!}{
\begin{tabular}{|c|c|c|c|c|c|}
\hline
Methods & c11 & c29 & c05 & c37 & average \\
\hline
Ours &\textbf{86.7} & \textbf{98.5} & 95.6 & \textbf{89.7} & \textbf{92.6} \\
\hline
CORAL~\cite{Sun_SFS_AAAI2016}  & 76.5 & 95.6 & 92.7 & 78.0 & 85.7 \\
\hline
SIDL \cite{Ni_NQC_CVPR2013}        & 76.5 & \textbf{98.5} & \textbf{98.5} & 88.2 & 90.4 \\
\hline
TJM \cite{Long_LWDSY_CVPR2014} & 83.8 & \textbf{98.5} & 95.6 & 82.4 & 90.1 \\
\hline
GFK \cite{Gong_GSSG_CVPR2012}            & 63.2 & 92.7 & 92.7 & 76.5 & 81.3  \\
\hline
Eigen light-field \cite{Gross_GMB_TPAMI2004}  & 78.0 & 91.0 & 93.0 & 89.0 & 87.8  \\
\hline
K-SVD \cite{Aharon_AEB_TSP2006}             & 48.5 & 76.5 & 80.9 & 57.4 & 65.8  \\
\hline
\end{tabular}}}
\end{table}

\textbf{Benefit of learning the common dictionary and domain-specific dictionaries separately}: Here we illustrate the benefits of separating the common dictionary and domain-specific dictionaries. Our method uses both the common dictionary and domain-specific dictionaries that are incoherent to the common dictionary to represent each intermediate domain while~\cite{Ni_NQC_CVPR2013} only uses a single dictionary to represent it. We want to compare the difference between the synthesized feature representations of the target data obtained by the above two methods. Therefore, we compute and visualize the synthesized faces of one target face in intermediate domains using our method and the method in~\cite{Ni_NQC_CVPR2013} as shown in Figure~\ref{fig::blur_face}. In addition, since our synthesized faces of the target face in the first two rows have two components corresponding to the common dictionary and domain-specific dictionaries respectively, we also visualize the two components in intermediate domains in the first two rows of Figure~\ref{fig::blur_face}. We observed that the synthesized faces obtained by two methods gradually transit from clear images to blur images. Moreover, the components that correspond to the common dictionary in the synthesized faces are always clear images while the components that correspond to the domain-specific dictionaries become more and more blurred. This shows that the common dictionary has the ability to exploit features shared by all the domains, and only the domain-specific dictionaries are used to exploit the domain shift. In addition, it can also be seen that improved reconstruction is achieved by our method, specifically for the region around the mouth where the motion blur is dominant. It further demonstrates that the separation of the common dictionary from domain-specific dictionaries enables us to learn more compact and discriminative representations for learning.

\subsubsection{Face Recognition Across Pose Variation}

We perform the second experiment on the CMU-PIE data set to evaluate the task of face recognition across pose variation. There are $5$ different poses of face images ranging from frontal to $\pm45^{o}$. The four non-frontal poses are denoted as \textit{c05} (yaw about $-22.5^{o}$), \textit{c29} (yaw about $22.5^{o}$), \textit{c11} (yaw about $45^{o}$) and \textit{c37} (yaw about $-45^{o}$). We select the front-illuminated face images to be labeled source domain. Face images with the same illumination condition under four non-frontal poses form the faces in the target domain.  The task is to classify the unlabeled face images in the target domain. Similar to face recognition across blur and illuminations, we compare our method with K-SVD~\cite{Aharon_AEB_TSP2006}, GFK~\cite{Gong_GSSG_CVPR2012}, TJM~\cite{Long_LWDSY_CVPR2014}, SIDL~\cite{Ni_NQC_CVPR2013} and CORAL~\cite{Sun_SFS_AAAI2016}. In addition, we compare with Eigen light-field~\cite{Gross_GMB_TPAMI2004}, which uses the appearance model to tackle pose variations in face recognition.

We report classification accuracies of different methods in Table~\ref{tab::pose}. As shown in Table~\ref{tab::pose}, our method outperforms its direct competitor~\cite{Ni_NQC_CVPR2013} under all cases except the case where the target pose is $c05$. It is interesting to note that when the pose variations are large,~\cite{Gross_GMB_TPAMI2004} which relies on a generic training set to build pose model has higher average recognition accuracies than the domain adaptation method proposed in~\cite{Gong_GSSG_CVPR2012}. However, our method demonstrates improved performances over both~\cite{Gross_GMB_TPAMI2004} and other domain adaptation approaches~\cite{Gong_GSSG_CVPR2012,Ni_NQC_CVPR2013,Long_LWDSY_CVPR2014,Sun_SFS_AAAI2016} when pose variations are large.

\begin{table*}[htp!]
\caption{{\bf Object classification accuracies of different approaches on the Office benchmark dataset} All $12$ pairs of combinations are evaluated with SURF features following the standard protocol of~\cite{Gong_GSSG_CVPR2012,Fernando_FHST_ICCV2013,Gopalan_GLC_ICCV2011,Saenko_SKFD_ECCV2010,Baktashmotlagh_BHLS_CVPR2014,Kulis_KSD_CVPR2011,Sun_SFS_AAAI2016}}
\label{tab::officeDA}
\centering{
\resizebox{0.96\textwidth}{!}{
\begin{tabular}{| c | c | c | c | c| c | c | c | c | c | c | c | c |}
\hline
Methods & C$\to$A & C$\to$D & C$\to$W & A$\to$C & A$\to$W & A$\to$D & W$\to$C & W$\to$A & W$\to$D & D$\to$C & D$\to$A & D$\to$W \\
\hline
NA & 43.7 & 39.4 & 30.0 & 35.8 & 24.9 & 33.1 & 25.7 & 32.3 & 78.9 & 27.1 & 26.4 & 56.4  \\ 
\hline 
K-SVD~\cite{Aharon_AEB_TSP2006} & 38.0 & 19.8 & 21.3& 33.9 & 23.5 & 22.3 & 17.1 & 16.7 & 46.5 & 22.6 & 14.3 & 46.8  \\
\hline
TCA~\cite{Pan_PTKY_IJCAI2009} & 46.7 & 41.4 & 36.2 & 40.0 & 40.1 & 39.1 & 33.7 & 40.2 & 77.5 & 34.0 & 39.6 & 80.4 \\ 
\hline 
GFK~\cite{Gong_GSSG_CVPR2012} &40.4& 41.1& 40.7& 37.9& 35.7& 36.3& 29.3& 35.5& 85.9& 30.3& 36.1& 79.1 \\
\hline
SVMA~\cite{Duan_DTX_TPAMI2012} & 39.1 & 34.5 & 32.9 & 34.8 & 32.5 & 34.1 & 33.5 & 36.6 & 75.0 & 31.4 & 33.4 & 74.4 \\
\hline 
SA~\cite{Fernando_FHST_ICCV2013} & 39.0 & 39.6& 23.9& 35.3& 38.6& 38.8&  32.3& 37.4& 77.8& 38.9& 38.0& 83.6 \\
\hline
SIDL~\cite{Ni_NQC_CVPR2013} & 43.3& 42.3& 36.3& 40.4& 37.9& 33.3& 36.3& 38.3& 86.2& 36.1& 39.1& 86.2 \\
\hline
TJM~\cite{Long_LWDSY_CVPR2014} & 46.7 & 44.6 & 38.9 & 39.4 & 42.0 & 45.2 & 30.2 & 30.0 & 89.2 & 31.4 & 32.8 & 85.4 \\
\hline
DIP~\cite{Baktashmotlagh_BHLS_ICCV13} & 50.0 & 49.0 & 47.6 & 43.3 & 46.7 & 42.8 & 37.0 & 42.5 & 86.4 & 39.0 & 40.5 & 86.7 \\
\hline
SIE~\cite{Baktashmotlagh_BHLS_CVPR2014} & 51.9 & 52.5 & 47.3 & \textbf{44.5} & \textbf{48.6} & 43.2 & 39.9 & \textbf{44.1} & 89.3 & 38.9 & 39.1 & 88.6 \\
\hline
CORAL~\cite{Sun_SFS_AAAI2016} & 47.2 & 40.7 & 39.2 & 40.3 & 38.7 & 38.3 & 34.6 & 37.8 & 84.9 & 34.2 & 38.1 & 85.9 \\ 
\hline 
Ours & \textbf{52.7} & \textbf{53.5} & \textbf{48.1} & 43.2 & 44.5 & \textbf{45.8} & \textbf{40.1} & 41.8 & \textbf{93.6} & \textbf{39.3} & \textbf{41.7} & \textbf{92.4} \\
\hline
\end{tabular}}}
\end{table*}

\textbf{Benefit of learning the common dictionary and domain-specific dictionaries separately}: In addition, we choose a target face with pose ID \textit{c11} and synthesized feature representations of this face images in intermediate domains. Since synthesized face images have two components corresponding to the common dictionary and domain-specific dictionaries respectively, we visualize the two components of synthesized face images separately in Figure~\ref{fig::pose_face}. It can be seen that the components corresponding to domain-specific dictionaries in intermediate domains gradually adapt from frontal face to non-frontal face. This demonstrates that the domain-specific dictionaries have the ability to encode the domain shift due to different yaw angles.

\subsection{Cross-domain Object Classification}


In this set of experiment, we evaluate our approach for cross-domain object classification using the standard Office benchmark dataset introduced in~\cite{Saenko_SKFD_ECCV2010}. This dataset contains visual objects across four different domains, \textit{i.e.} Caltech, Amazon, DSLR, Webcam.The Caltech~\cite{Griffin_GHP_Tech2007} domain contains $256$ object classes with images from Google. The Amazon domain consists of $31$ classes with different image instances downloaded from Amazon website. Images from both Caltech and Amazon have large intra-class variations. The DLSR domain includes images acquired from a digital SLR (DSLR) camera with also total $31$ categories. The last domain Webcam contains images from a webcam with similar environment as the DSLR. However, the images from Webcam have much lower resolution with significant noise. Following the standard protocol as in~\cite{Gong_GSSG_CVPR2012,Fernando_FHST_ICCV2013,Gopalan_GLC_ICCV2011,Saenko_SKFD_ECCV2010,Baktashmotlagh_BHLS_CVPR2014,Kulis_KSD_CVPR2011,Sun_SFS_AAAI2016}, we select $10$ object classes ( backpack, touring bike, calculator, headphones, computer keyboard, computer monitor, computer mouse, coffee mug, video projector, laptop $101$) common to all four domains with a total of $2,533$ images for our experiments. Each class in each domain contains $8$ to $151$ images. 

\subsubsection{Object Classification with Shallow Features} 

We follow the standard protocol in~\cite{Gong_GSSG_CVPR2012,Fernando_FHST_ICCV2013,Gopalan_GLC_ICCV2011,Saenko_SKFD_ECCV2010,Baktashmotlagh_BHLS_CVPR2014,Kulis_KSD_CVPR2011,Sun_SFS_AAAI2016} and evaluate the proposed approach with shallow features. Specifically, all the images were firstly resized to have the same width and converted to grayscale. Second, the SURF detector~\cite{Bay_BETV_CVIU2008} was used to extract local scale-invariant interest points. Then a random subset of these interest point descriptors were quantized to $800$ visual words by $k$-means clustering. Each image was finally represented by a $800$-dimensional histogram. Note that there are total $12$ experiment settings from the combination of four domains. These are C$\to$A (Source domain is (C)altech and target domain is (A)mazon), C$\to$D (train on (C)altech and test on (D)SLR), and so on. Following the standard settings~\cite{Gong_GSSG_CVPR2012,Fernando_FHST_ICCV2013,Gopalan_GLC_ICCV2011,Saenko_SKFD_ECCV2010,Baktashmotlagh_BHLS_CVPR2014,Kulis_KSD_CVPR2011,Sun_SFS_AAAI2016}, in the source domain, we randomly selected $20$ labeled images per category when Amazon, Webcam and Caltech are used as source domains, and $8$ labeled images when DSLR is the source domain. Moreover, we run $20$ different trials corresponding to different selections of labeled source data and report the average recognition accuracy in Table~\ref{tab::officeDA}.

In Table~\ref{tab::officeDA}, we compare our method with several baseline and recent published methods: Baseline 1 (NA). No adaptation baseline. Directly train a linear SVM from source domain and test on the target domain. Baseline 2 (K-SVD)~\cite{Aharon_AEB_TSP2006}. We learn dictionaries using K-SVD in~\cite{Aharon_AEB_TSP2006} and decomposes the test samples with dictionaries. We classify the test sample using nearest neighbor classifier on the sparse codes. TCA~\cite{Pan_PTKY_IJCAI2009}, GFK~\cite{Gong_GSSG_CVPR2012}, SVMA~\cite{Duan_DTX_TPAMI2012}, SA~\cite{Fernando_FHST_ICCV2013}, SIDL \cite{Ni_NQC_CVPR2013}, TJM \cite{Long_LWDSY_CVPR2014}, DIP~\cite{Baktashmotlagh_BHLS_ICCV13}, SIE~\cite{Baktashmotlagh_BHLS_CVPR2014} and CORAL~\cite{Sun_SFS_AAAI2016}. GFK, SA, TCA and SIE are manifold-base methods that projected the source and target distributions into a low-dimensional manifold. In detail, SA~\cite{Fernando_FHST_ICCV2013} aligned the source and target subspaces by finding a linear mapping that minimizes the Frobenius norm of the difference. SIE~\cite{Baktashmotlagh_BHLS_CVPR2014} extended this by measuring the Hellinger distance on the statistical manifolds. CORAL~\cite{Sun_SFS_AAAI2016} minimized the domain shift by aligning the second-order statistics of source and target distribution. It can be seen that our method achieve the best performance for a majority of combinations of source and target domains. In particular, our method consistently outperforms SIDL \cite{Ni_NQC_CVPR2013} which is most similar to ours. This is because~\cite{Ni_NQC_CVPR2013} only regularizes two adjacent domains to have the identical pairwise sparse codes and the learned dictionaries do not fully capture the domain changes. However, our method encodes the domain changes in the domain-specific dictionaries by encouraging feature representation of different domain data to have the same domain-adaptive sparse codes.


\begin{table}[htp]
\caption{{\bf Object classification accuracies of different approaches on the Office benchmark dataset} All $6$ pairs of combinations are evaluated with deep features following the standard protocol of~\cite{Donahue_DJVHZTD_ICML2014,Tzeng_THZSD_2014,Long_LCWJ_ICML2015,Ganin_GL_ICML15,Sun_SFS_AAAI2016}}
\label{tab::office_Deep}
\centering{
\resizebox{0.49\textwidth}{!}{
\begin{tabular}{| c | c | c | c | c | c | c | }
\hline
Methods &  A $\to$ D & A $\to$ W & D $\to$ A & D $\to$ W & W $\to$ A & W $\to$ D \\
\hline
NN-fc7 & 55.7 & 50.6 & 46.5 & 93.1 & 43.0 & 97.4 \\
\hline
NN-FT7 & 58.5 & 53.0 & 43.8 & 94.8 & 43.7 & 99.1 \\
\hline
\hline
TCA-fc7 & 45.4 & 40.5 & 36.5 & 78.2 & 34.1 & 85 \\
\hline
TCA-FT7 & 47.3 & 45.2 & 36.4 & 80.9 & 39.2 & 92 \\
\hline
GFK-fc7 & 52 & 48.2 & 41.8 & 86.5 & 38.6 & 87.5 \\
\hline
GFK-FT7 & 56.4 & 52.3 & 43.2 & 92.2 & 41.5 & 96.6 \\
\hline
SA-ft7 & 46.2 & 42.5 & 39.3 & 78.9 & 36.3 & 80.6 \\
\hline
SA-FT7 & 50.5 & 47.2 & 39.6 & 89.0 & 37.3 & 93.0 \\ 
\hline
CORAL-fc7 & 57.1 & 53.1 & 51.1 & 94.6 & 47.3 & 98.2 \\
\hline
CORAL-FT7 & 62.2 & 61.9 & 48.4 & 96.2 & 48.2 & \textbf{99.5}  \\
\hline
\hline
DLID~\cite{Chopra_CBG_ICMLW2013} & - & 26.1 & - & 68.9 & - & 84.9 \\
\hline
DANN~\cite{Ghifary_GKZ_PRICAI2014} & 34.0 & 34.1 & 20.1 & 62.0 & 21.2 & 64.4 \\
\hline
DECAF-fc7~\cite{Donahue_DJVHZTD_ICML2014} & - & 53.9 & - & 89.2 & - & - \\
\hline 
DDC~\cite{Long_LCWJ_ICML2015} & - & 59.4 & - & 92.5 & - & 91.7 \\
\hline
ReverseGrad~\cite{Ganin_GL_ICML15} & - & \textbf{67.3} & - & 94.0 & - & 93.7 \\
\hline
\hline
DADL-fc6 & 54.5 & 50.1 & 47.3 & 95.6 & 42.2 & 97.9 \\ 
\hline
DADL-fc7 & 59.3 & 54.5 & 51.2 & 95.3 & 47.1 & 98.2 \\
\hline
DADL-FT6 & 63.2 & 60.7 & 50.6 & 97.1 & 46.7 & \textbf{99.5} \\
\hline
DADL-FT7 & \textbf{64.0} & 62.4 & \textbf{52.0} & \textbf{98.2} & \textbf{48.4} & \textbf{99.5} \\
\hline
\end{tabular}}
}
\end{table}

\subsubsection{Object Classification with Deep Features} 

In the second part, we evaluate the proposed DADL approach and carry out the experiments on the standard Office dataset with deep features. In order to have fair comparison, we evaluate the our method using both the pre-trained AlexNet~\cite{Krizhevsky_KSH_NIPS2012} (DADL-fc6, DADL-fc7) and AlexNet fine-tuned on the source (DADL-FT6 and DADL-FT7), following the standard protocol in~\cite{Donahue_DJVHZTD_ICML2014,Tzeng_THZSD_2014,Long_LCWJ_ICML2015,Ganin_GL_ICML15,Sun_SFS_AAAI2016}. The baseline results are obtained as the fine-tuning only on the source domain (NN-FT6 and NN-FT7). Note that in this setting, there are three domains, which results in total $6$ combination of source and target domains. We run $5$ different trails corresponding to the different splits of source and target data and report the mean accuracy in Table~\ref{tab::office_Deep}.

\begin{table*}[t]
\caption{{\bf Recognition accuracy of face component across different sessions on the UMDAA-01 dataset \cite{Zhang_ZPFC_WACV2015}.}}
\label{tab::UMDAA_face}
\centering{
\resizebox{0.6\textwidth}{!}{
\begin{tabular}{| c | c | c | c | c | c | c | c |}
\hline
Methods & $1 \to 2$ & $1 \to 3$ & $2 \to 3$ & $2 \to 1$ & $3 \to 1$ & $3 \to 2$ & Average  \\
\hline
Ours      & \textbf{50.24} & \textbf{60.67} & \textbf{65.60} & \textbf{55.68} & \textbf{59.70} & \textbf{60.34} & \textbf{58.71} \\
\hline
SIDL \cite{Ni_NQC_CVPR2013}             & 44.08 & 56.64 & 61.54 & 50.08 & 55.77 & 56.24 & 54.05 \\
\hline
SA \cite{Fernando_FHST_ICCV2013}     & 40.92 & 49.22 & 51.84 & 46.32 & 51.91 & 47.86 & 48.01 \\
\hline
K-SVD \cite{Aharon_AEB_TSP2006}        & 34.80 & 44.27 & 51.70 & 44.47 & 47.40 & 42.53 & 44.20 \\
\hline
NN                                   & 39.33 & 50.53 & 53.60 & 44.20 & 51.13 & 47.13 & 47.65 \\
\hline
\end{tabular}}
}
\end{table*}

\begin{table*}[h]
\caption{{\bf Recognition accuracy of touch gesture component across different sessions on the UMDAA-01 dataset \cite{Zhang_ZPFC_WACV2015}.}}
\label{tab::UMDAA_touch}
\centering{
\resizebox{0.6\textwidth}{!}{
\begin{tabular}{| c | c | c | c | c | c | c | c |}
\hline
Methods & $1 \to 2$ & $1 \to 3$ & $2 \to 3$ & $2 \to 1$ & $3 \to 1$ & $3 \to 2$ & Average  \\
\hline
Ours      & \textbf{29.29} & \textbf{29.70} & \textbf{34.60} & \textbf{27.55} & \textbf{28.60} & \textbf{32.97} & \textbf{30.46} \\
\hline
SIDL \cite{Ni_NQC_CVPR2013}             & 23.76 & 24.08 & 28.36 & 22.93 & 24.14 & 26.75 & 25.00 \\
\hline
SA \cite{Fernando_FHST_ICCV2013}     & 23.02 & 22.61 & 26.21 & 22.12 & 24.63 & 27.98 & 23.01 \\
\hline
K-SVD \cite{Aharon_AEB_TSP2006}        & 14.72 & 15.44 & 17.52 & 14.43 & 16.96 & 19.49 & 16.43 \\
\hline
NN                                   & 20.75 & 21.22 & 23.59 & 19.48 & 22.49 & 24.35 & 21.98 \\
\hline
\end{tabular}}
}
\end{table*}

Besides TCA~\cite{Pan_PTKY_IJCAI2009}, GFk~\cite{Gong_GSSG_CVPR2012}, SA~\cite{Fernando_FHST_ICCV2013} and CORAL~\cite{Sun_SFS_AAAI2016}, we also compare the proposed method with other state-of-the-art deep learning approaches, as shown in Table~\ref{tab::office_Deep}:
\begin{itemize}
\item DLID~\cite{Chopra_CBG_ICMLW2013} trained a CNN architecture jointly on source and target domain with n interpolating path. 
\item DANN~\cite{Ghifary_GKZ_PRICAI2014} reduced the domain distribution shift by incorporating the MMD measurement as regularization.
\item DECAF~\cite{Donahue_DJVHZTD_ICML2014} applied AlexNet~\cite{Krizhevsky_KSH_NIPS2012} pre-tarined on the ImageNet and extracted features from the fc6 and fc7 layers on the source domains to train a classifier. Target samples were applied by using the classifier directly trained from the source domain. 
\item DDC~\cite{Tzeng_THZSD_2014} added a domain shifting loss to AlexNet~\cite{Krizhevsky_KSH_NIPS2012} and further fine-tuned on both the source and target domain. 
\item DAN~\cite{Long_LCWJ_ICML2015} utilized a multi-kernel selection method for better adapted embedding in multiple layers
\item ReverseGrad~\cite{Ganin_GL_ICML15} learned domain-invariant features by treating the source and target domain as two classes and adding a bi-classfication task. 
\end{itemize}

It can be seen that our method achieve the best performance for a majority of combinations of source and target domains on the deep features. Note that some deep architecture based methods only reported results on few settings. Since higher layer ft7/FT7 features lead to superior performance over the lower layer ft6/FT6, we only listed results from ft7 and FT7 of TCA~\cite{Pan_PTKY_IJCAI2009}, GFK~\cite{Gong_GSSG_CVPR2012}, SA~\cite{Fernando_FHST_ICCV2013} and CORAL~\cite{Sun_SFS_AAAI2016} here. Moreover, in some cases, the NN baseline method achieve even better performance than the manifold-based methods and some deep methods. However, our method consistently outperforms the baseline performance. 

\subsection{Cross-domain Active Authentication}

\begin{figure}[h]
 \centering
 \includegraphics[width=6cm]{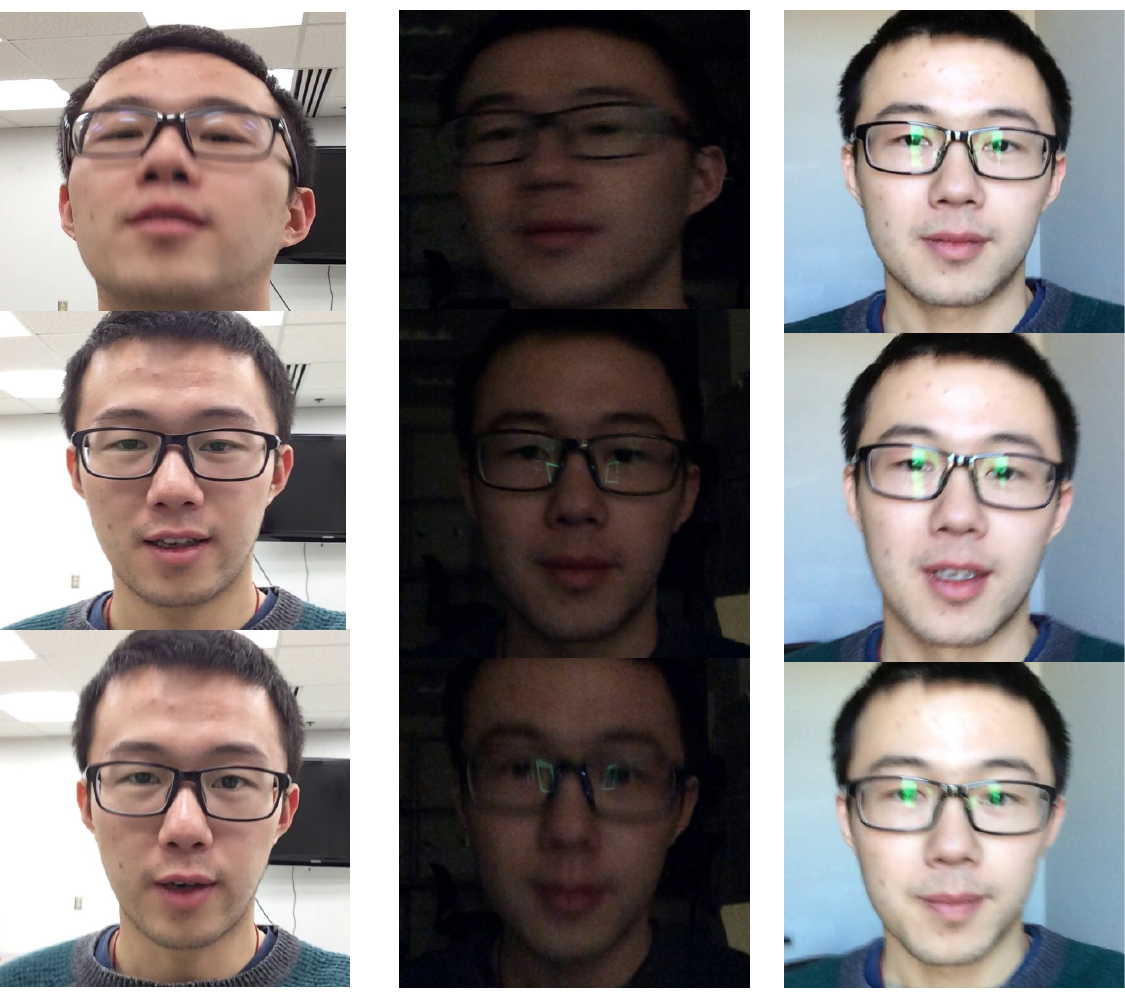}\\
 (a)\hskip50pt(b)\hskip50pt(c)
 \caption{Samples images from the UMDAA-01 dataset.  (a), (b) and (c) show some samples images from session 1, 2 and 3, respectively.}
\label{fig:umddata}
\end{figure}

In this section, we evaluated our approach for face and gesture recognition using the UMD Active Authentication dataset (UMDAA-$01$) dataset first introduced in~\cite{Zhang_ZPFC_WACV2015}. UMDAA-$01$ dataset~\cite{Zhang_ZPFC_WACV2015} is a new challenging dataset, which contains data of both face images and touch gestures collected from $50$ different mobile users. Each user was asked to perform $5$ different task: enrollment task, document task, picture task, popup task and scrolling task in $3$ different sessions with different illumination conditions. The touch gesture data and face image data were collected simultaneously, which resulted in $15,490$ touch gestures and $750$ videos consisting of facial data. In addition, since all the facial data was collected in an unconstrained manner, the faces also exhibited different poses and blur variations. In summary, the domain shift consists of poses, blur and illumination variations across different sessions. Figure~\ref{fig:umddata} shows some sample images from the UMDAA-$01$ dataset.

We follow the standard protocol in~\cite{Zhang_ZPSC_FG2015,Zhang_ZPFC_WACV2015} and evaluate the proposed approach on two modalities (1) Faces component and (2) Touch Gestures Component. Specifically, for the face data, landmarks of the face images were detected in each frame of the videos using face landmarks detector as in~\cite{Zhu_ZR_CVPR2012}. The cropped and aligned face images were further rescaled to the size of $192 \times 168 \times 3$.  All the face images were converted to grayscale and illumination normalization as further applied. Finally, each image was represented by a $224$-dimensional vector from the down sampling the images to $16 \times 14$.
For the touch gestures, a $27$-dimensional feature vector was extracted for every single swipe using the method in~\cite{Zhang_ZPFC_WACV2015}. Because different sessions have different underlying characteristics of the data collection, recognizing the data across different sessions could be viewed as cross-domain recognition problem. Note that there are total $6$ experiment settings from the combination of three different sessions ($1 \to 2$, $1 \to 3$, $2 \to 3$, $2 \to 1$, $3 \to 1$, $3 \to 2$). For the face component, we selected $30$ faces from each subject in each session, which results in a total of $4,500$ face images of $50$ users across $3$ different sessions. For the touch gestures component, we selected same $4,500$ touch swipes of $50$ users across $3$ sessions. We select $20$ samples for each user from one session as training data to form the source domain. The data samples from the other session are used as test data. Moreover, we run each experiment $10$ times and report the average recognition accuracy. 

\begin{figure*}[t]
\centering{
\begin{subfigure}{.24\textwidth}
  \centering
  \includegraphics[width = 1.8in,height = 1.3in]{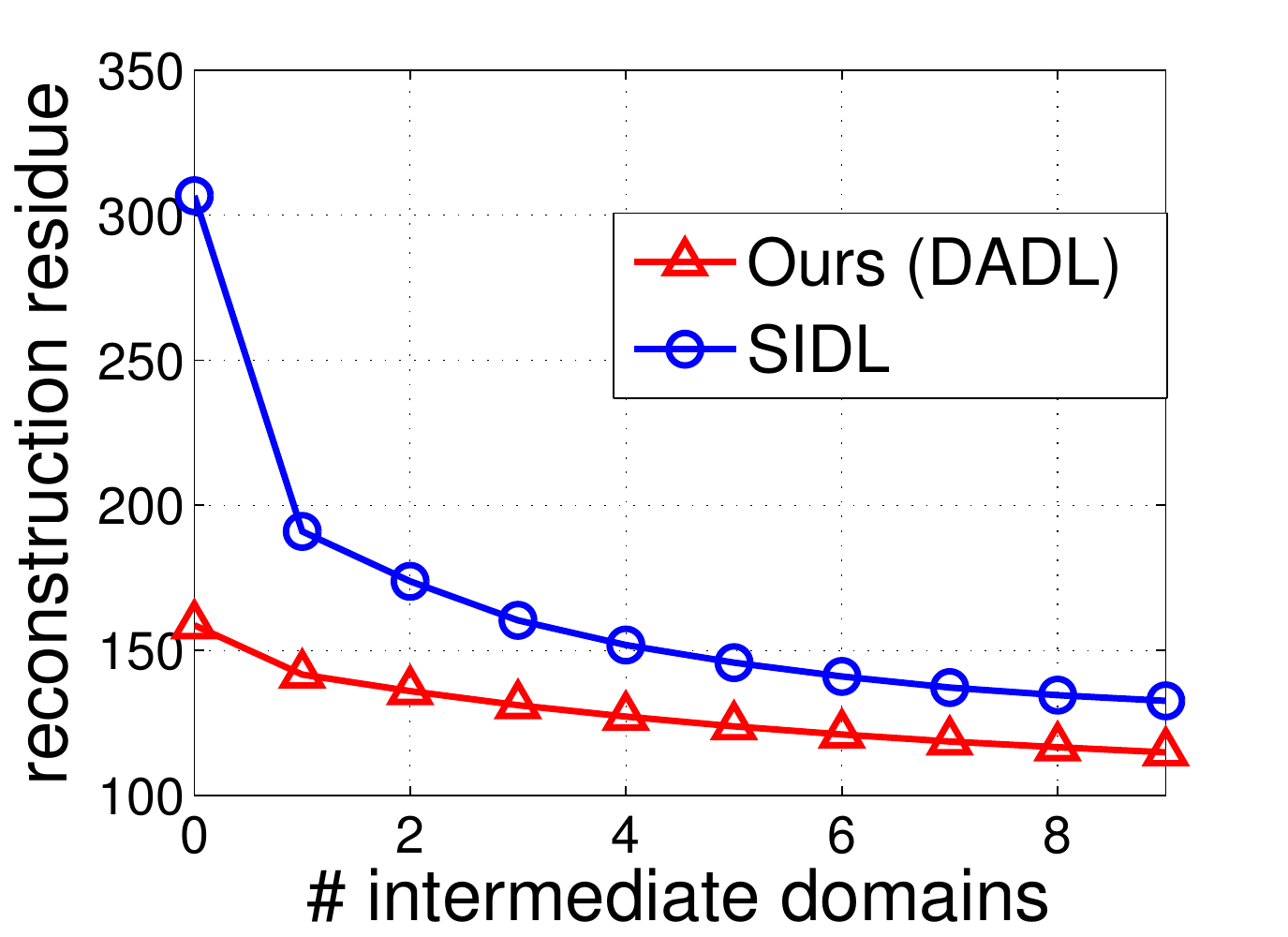}
\end{subfigure}%
\begin{subfigure}{.24\textwidth}
  \centering
  \includegraphics[width = 1.8in,height = 1.3in]{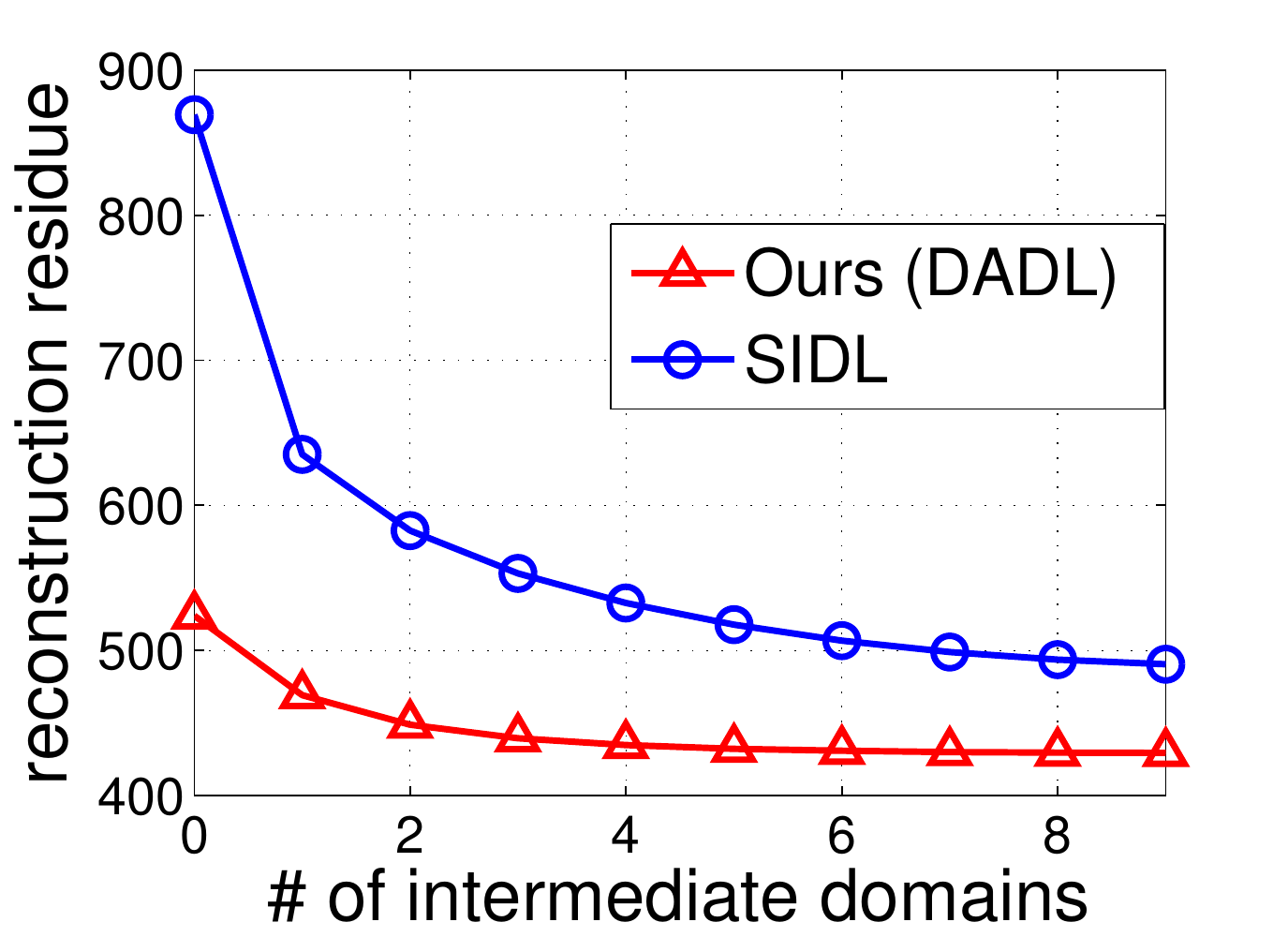}
\end{subfigure}%
\begin{subfigure}{.24\textwidth}
  \centering
  \includegraphics[width = 1.8in,height = 1.3in]{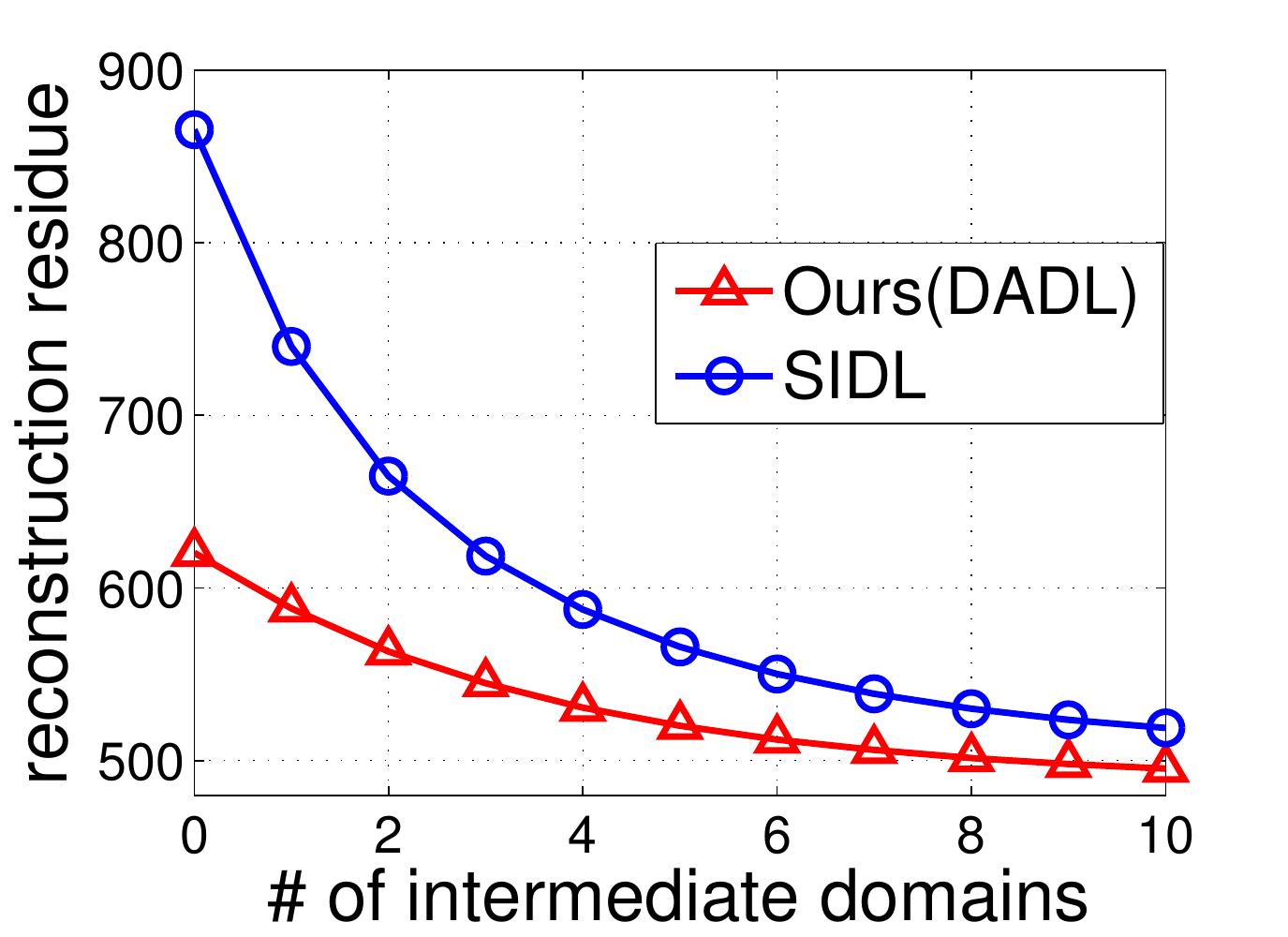}
\end{subfigure}%
\begin{subfigure}{.24\textwidth}
  \centering
  \includegraphics[width = 1.8in,height = 1.3in]{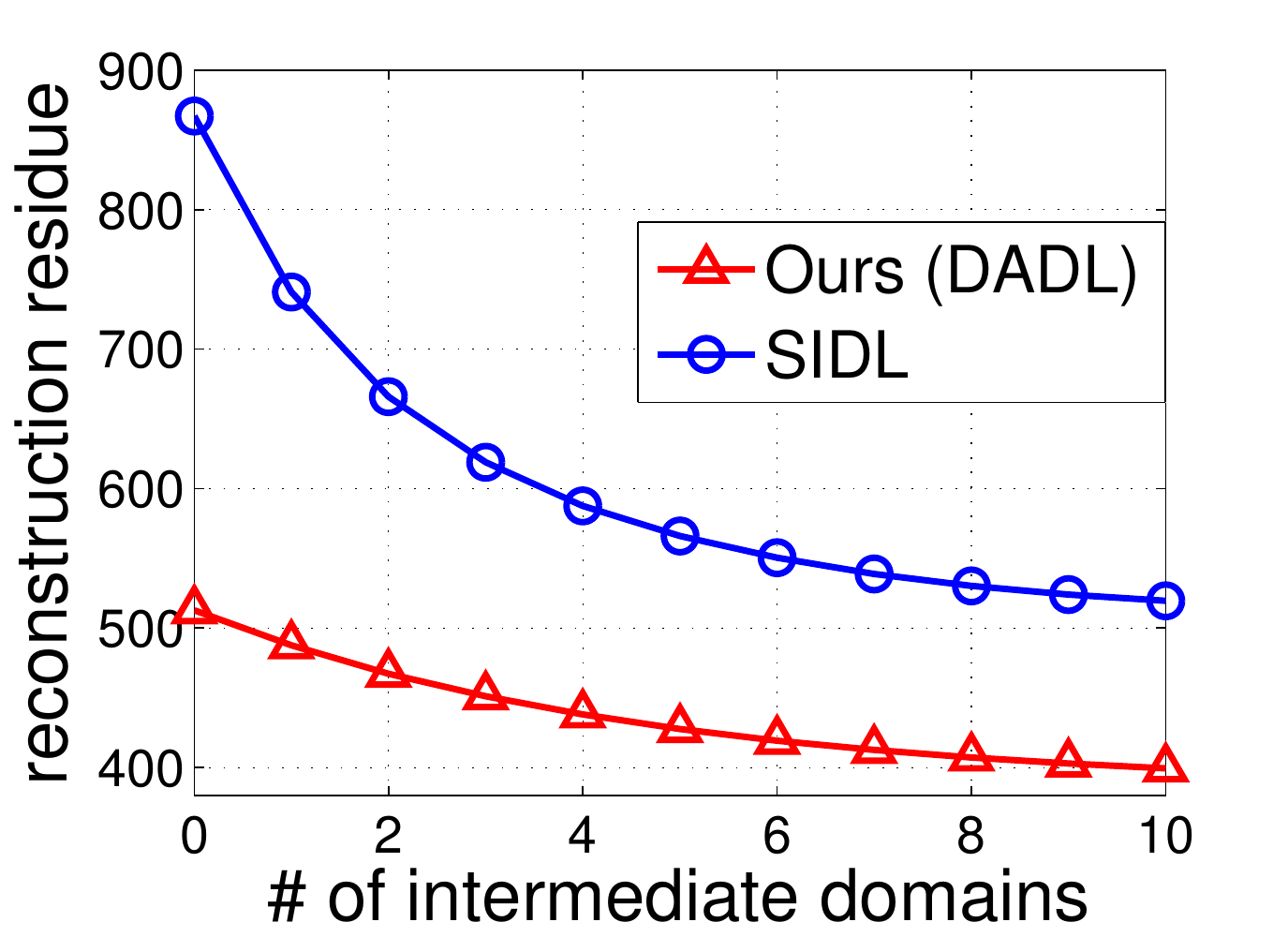}
\end{subfigure}%
}
\caption{{\bf Average reconstruction error of the target data decomposed using dictionaries along the intermediate domains.} The source and target domains are: (a) clear frontal face v.s. motion blur frontal face ($L=5$) (b) frontal face images v.s. face images at pose $c05$ (c) Amazon v.s. Caltech in Office dataset~\cite{Saenko_SKFD_ECCV2010} (d) Webcam v.s. Amazon in Office dataset~\cite{Saenko_SKFD_ECCV2010}. We also compare our method with SIDL~\cite{Ni_NQC_CVPR2013}}
\label{fig::decrease_residue}
\end{figure*}

\begin{figure*}[htp]
\begin{subfigure}{.33\textwidth}
  \centering
  \includegraphics[width = 2.1in,height = 1.5in]{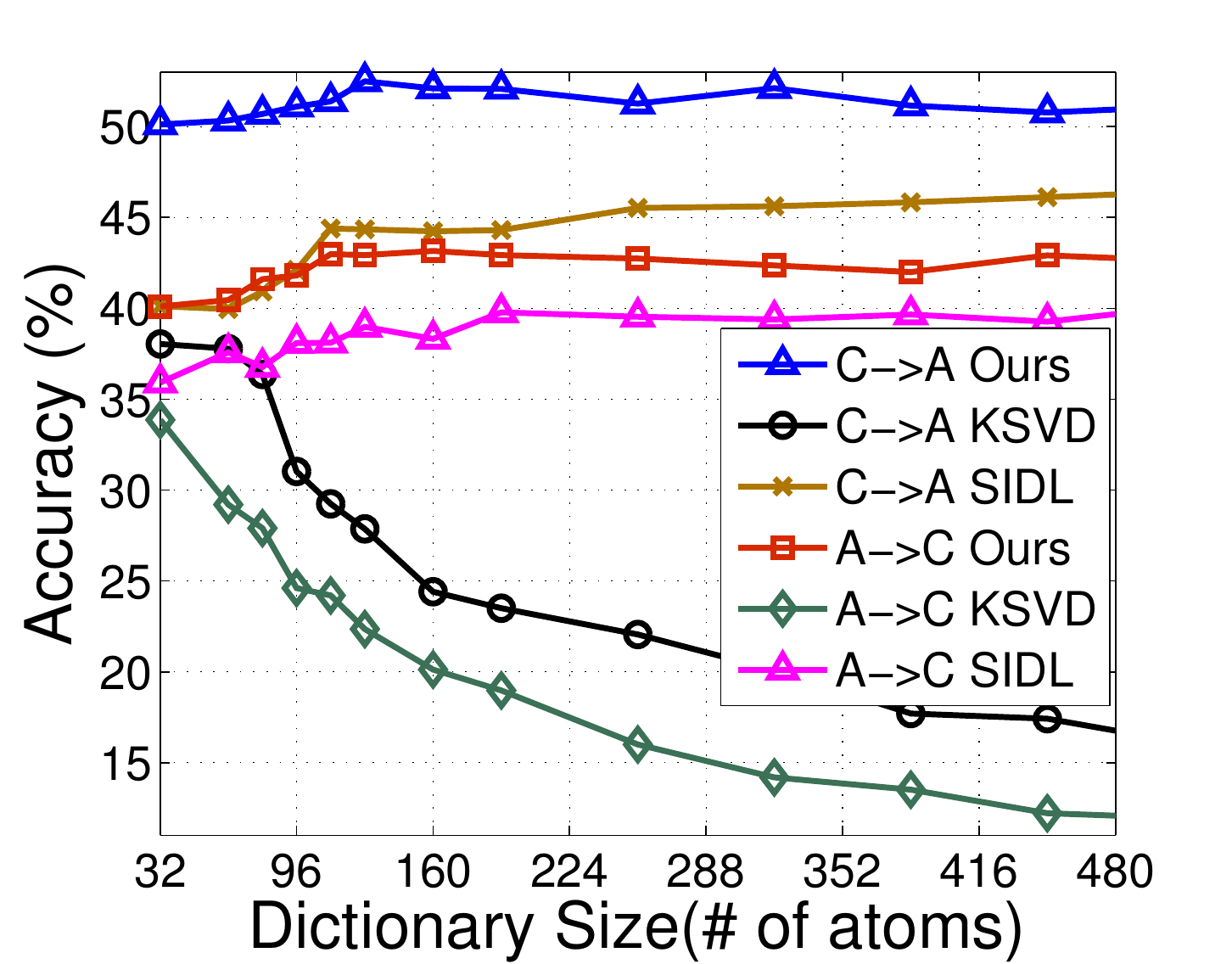}
  \caption{dictionary size}
  \label{fig:OFFICE_dictionary size}
\end{subfigure}%
\begin{subfigure}{.33\textwidth}
  \centering
  \includegraphics[width = 2.1in,height = 1.5in]{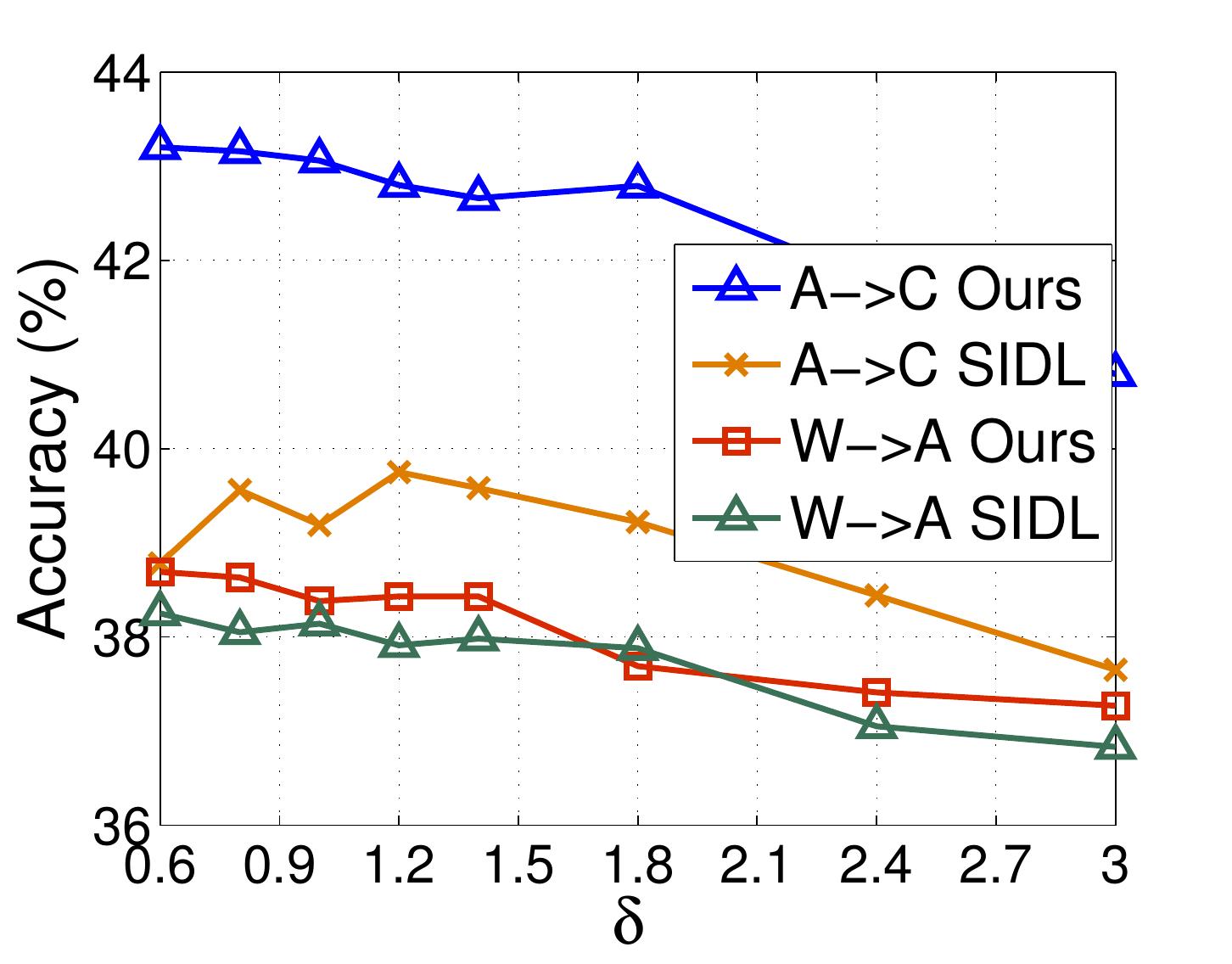}
  \caption{stopping threshold $\delta$}
  \label{fig:OFFICE_delta}
\end{subfigure}%
\begin{subfigure}{.33\textwidth}
  \centering
  \includegraphics[width = 2.1in,height = 1.5in]{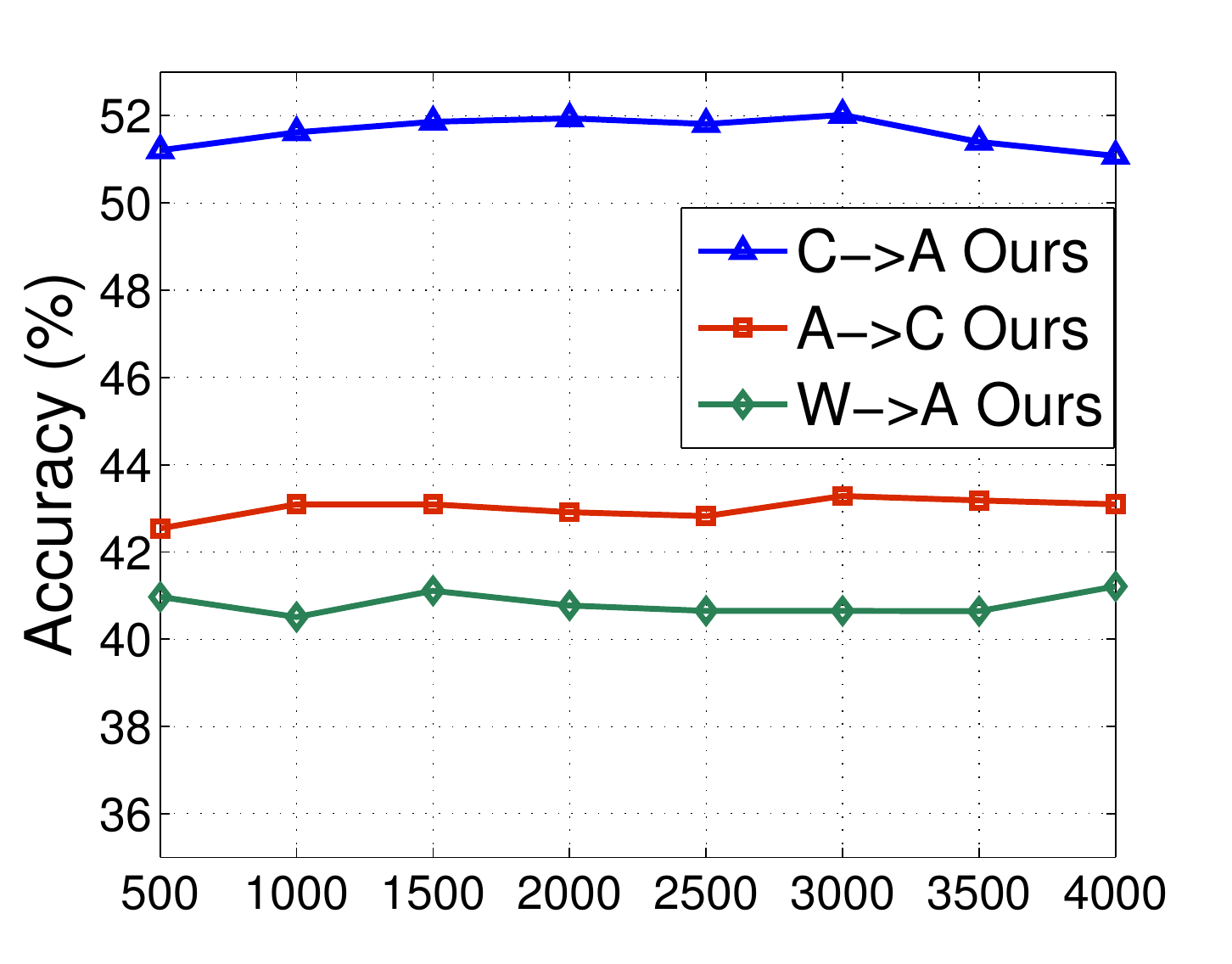}
  \caption{regularization parameter $\eta$}
  \label{fig:OFFICE_eta}
\end{subfigure}%
\caption{Parameter Sensitivity: effect of (a) dictionary size, and (b) stopping threshold $\delta$ and (c) regularization parameter $\eta$ on Office dataset~\cite{Saenko_SKFD_ECCV2010}}
\label{fig::parameter_sensitivity}
\end{figure*}

We compare our method with several baseline and other domain adaptation methods for both face and touch gesture data. The average accuracy is reported in Table~\ref{tab::UMDAA_face} and Table~\ref{tab::UMDAA_touch}, respectively. It can be seen that the  our method outperforms the other methods on all $6$ domain pairs, sometimes by a large margin. In particular, our method also consistently outperforms SIDL~\cite{Ni_NQC_CVPR2013}, which is most similar to ours in spirit. Moreover, the performance of the touch gesture component degenerates significantly compared with the face component. This is due to the large diversity and variations of the swipe gesture features. How to extract the discriminative and robust features from swipe gesture is one of the interesting topics for the active authentication problem. 

%

\subsection{Empirical Analysis}
\label{sec::empirical_analysis}
In this part, we go deeper into the proposed DADL method by investigating the reconstruction error and parameter sensitivity to further show the effectiveness of our method. 

\subsubsection{Reconstruction Error}
One advantage of the proposed method is that the source-specific dictionary gradually adapts to the target domain-specific dictionary through a set of intermediate domain-specific dictionaries.
Thus we are curious to find how the average reconstruction error in equation~(\ref{eqn::residue target}), using both the common dictionary and domain-specific dictionaries, changes along the transition path we have learned. 

In Figure~\ref{fig::decrease_residue}, we observe that reconstruction residues obtained by our method are gradually decreasing along the transition path, which provides empirical support to Proposition~\ref{prop::1}. Moreover, our method achieves much lower reconstruction error compared with SIDL~\cite{Ni_NQC_CVPR2013}. This demonstrates that our approach can learn more compact and more reconstructive dictionaries by learning the common and domain-specific dictionaries separately. Last but not the least, our learning algorithm generally terminates within $8$ to $10$ steps, which demonstrates that the generated intermediate domains bridge the gap between the source and target domains. 


\subsubsection{Parameter Sensitivity}

We conduct empirical parameter sensitivity analysis, which validates the proposed DADL method could achieve promising performance under wide range of parameter values. In order to evaluate the effect of dictionary size on our approach, we choose two different combinations of source and target domains and plot the results in Figure~\ref{fig:OFFICE_dictionary size}. We also evaluate our approach with varying values of stopping threshold $\delta$ as shown in Figure~\ref{fig:OFFICE_delta}. In addition, we plot the results of recognition accuracy in Figure~\ref{fig:OFFICE_eta}  with different values of regulization parameter $\eta$ in~\eqref{eqn::objective_deltaD}.

It can been seen that in Figure~\ref{fig:OFFICE_dictionary size}, our approach yields significant improvement over K-SVD~\cite{Aharon_AEB_TSP2006} since we bridge the domain shift by generating intermediate domains. Our approach also outperforms SIDL~\cite{Ni_NQC_CVPR2013} by a large margin of $4.5\%$. This is because we learn more compact and reconstructive dictionaries to represent target data, which leads to much lower reconstruction errors, as demonstrated in Figure~\ref{fig::decrease_residue}. The dictionary size is set to be $128$ or $256$ based on the source sample size in all the experiments. It can be seen that both~\cite{Ni_NQC_CVPR2013} and the proposed approach converge in fewer steps with increasing value of $\delta$, thus generating fewer intermediate domains. In addition, our approach is insensitive to the regulization parameter $\eta$, which is chosen from $\left[ 1500, 2000, 2500 \right]$ throughout all the experiments. 


\section{Conclusion}
\label{sec::conclusion}
We presented a novel domain adaptive dictionary learning framework for cross-domain visual recognition. We first learned a common dictionary to recover features shared by all domains. Then we acquired a set of domain-specific dictionaries, which generates a transition path from source to target domains. The common dictionary is essential for reconstruction while domain-specific dictionaries are able to bridge the domain shift. Final feature representations are recovered by utilizing both common and domain-specific dictionaries. We extensively evaluated our approach on three benchmark datasets for cross-domain visual recognition and the experimental results clearly confirmed the effectiveness of our approach.

\appendix

\noindent In this section, we provide the proof of Proposition~\ref{prop::1}.

\noindent Proof: Define the Singular Value Decomposition (SVD) of $\mathbf{\Gamma}^{k} \in \mathbb{R}^{n\times N_t}$ as $\mathbf{\Gamma}^{k}=\mathbf{U}\mathbf{\Lambda}\mathbf{V}^{T}$, where $\mathbf{U}\mathbf{U}^T = \mathbf{I}\in\mathbb{R}^{n\times n}$ and $\mathbf{V}\mathbf{V}^{T} = \mathbf{I} \in\mathbb{R}^{N_t\times N_t} $, and $\mathbf{\Lambda}=[\tilde{\mathbf{\Lambda}}, \mathbf{0}] \in \mathbb{R}^{n\times N_t}$ is a rectangular diagonal matrix, with $\tilde{\mathbf{\Lambda}}=diag(\lambda_{1},...,\lambda_{n})$ being a diagonal matrix. We decompose $\mathbf{V}=[\mathbf{V}_{1}, \mathbf{V}_{2}]$ where $\mathbf{V}_{1} \in\mathbb{R}^{N_t\times n}$ and evaluate the following term below:
\begin{align}
&{\mathbf{\Gamma}^{k}}^{T}(\eta\mathbf{I}+\mathbf{\Gamma}^{k}{\mathbf{\Gamma}^{k}}^{T})^{-1}\mathbf{\Gamma}^{k} \notag \\
= & \mathbf{V}\mathbf{\Lambda}^{T}\mathbf{U}^{T}(\eta\mathbf{I}+\mathbf{U}\mathbf{\Lambda}\mathbf{\Lambda}^{T}\mathbf{U}^{T})^{-1}\mathbf{U\Lambda}\mathbf{V}^{T} \notag \\
= & [\mathbf{V}_{1},\mathbf{V}_{2}]\mathbf{\Lambda}^{T}\mathbf{U}^{T}(\eta\mathbf{I}+\mathbf{U}\tilde{\mathbf{\Lambda}}^{2}\mathbf{U}^{T})^{-1}\mathbf{U}\mathbf{\Lambda}[\mathbf{V}_{1},\mathbf{V}_{2}]^{T} \notag \\
= & [\mathbf{V}_{1},\mathbf{V}_{2}]\mathbf{\Lambda}^{T}\mathbf{U}^{T}(\mathbf{U}(\eta\mathbf{I})\mathbf{U}^{T}+\mathbf{U}\tilde{\mathbf{\Lambda}}^{2}\mathbf{U}^{T})^{-1}\mathbf{U}\mathbf{\Lambda}[\mathbf{V}_{1},\mathbf{V}_{2}]^{T} \notag \\
= & \mathbf{V}_{1}\tilde{\mathbf{\Lambda}}(\eta\mathbf{I}+\tilde{\mathbf{\Lambda}}^{2})^{-1}\tilde{\mathbf{\Lambda}}\mathbf{V}_{1}^{T} \label{eq:sparsecode_svd} \\
= & \mathbf{V}_{1}\mathbf{\Theta}\mathbf{V}_{1}^{T} \notag
\end{align}
where $\mathbf{\Theta}=diag(\mathbf{\overrightarrow{\theta}}) = diag(\frac{\lambda_{1}^{2}}{\lambda_{1}^{2}+\eta},...,\frac{\lambda_{n}^{2}}{\lambda_{n}^{2}+\eta})$.

Next, we replace $\Delta \mathbf{D}^{k}$ with~\eqref{eq::dictionary_update_1} and have
\begin{align}
& \|\mathbf{J}^{k} - \Delta\mathbf{D}^{k}\mathbf{\Gamma}^{k}\|_{F}^{2} - \|\mathbf{J}^{k}\|_{F}^{2} \notag\\
 = &\|\mathbf{J}^{k} - \mathbf{J}^{k}{\mathbf{\Gamma}^{k}}^{T}(\eta\mathbf{I}+\mathbf{\Gamma}^{k}{\mathbf{\Gamma}^{k}}^{T})^{-1}\mathbf{\Gamma}^{k}\|_{F}^{2} - \|\mathbf{J}^{k}\|_{F}^{2}  \label{eq::prop1_expand} \\
 = & tr(\mathbf{J}^{k}{\mathbf{J}^{k}}^{T}) - tr(2{\mathbf{\Gamma}^{k}}^{T}(\eta\mathbf{I}+\mathbf{\Gamma}^{k}{\mathbf{\Gamma}^{k}}^{T})^{-1}{\mathbf{\Gamma}^{k}}{\mathbf{J}^{k}}^{T}\mathbf{J}^{k}) \notag \\
 + & tr({\mathbf{\Gamma}^{k}}^{T}(\eta\mathbf{I}+\mathbf{\Gamma}^{k}{\mathbf{\Gamma}^{k}}^{T})^{-1}\mathbf{\Gamma}^{k}{\mathbf{J}^{k}}^{T}\mathbf{J}^{k}{\mathbf{\Gamma}^{k}}^{T}(\eta\mathbf{I}+{\mathbf{\Gamma}^{k}}^{T}\mathbf{\Gamma}^{k})^{-1}\mathbf{\Gamma}^{k}) \notag\\
 - & tr(\mathbf{J}^{k}{\mathbf{J}^{k}}^{T}) \notag \\
 = & tr({\mathbf{\Gamma}^{k}}^{T}(\eta\mathbf{I}+\mathbf{\Gamma}^{k}{\mathbf{\Gamma}^{k}}^{T})^{-1}\mathbf{\Gamma}^{k}{\mathbf{J}^{k}}^{T}\mathbf{J}^{k}{\mathbf{\Gamma}^{k}}^{T}(\eta\mathbf{I}+{\mathbf{\Gamma}^{k}}^{T}\mathbf{\Gamma}^{k})^{-1}\mathbf{\Gamma}^{k}) \notag \\
 & - tr(2{\mathbf{\Gamma}^{k}}^{T}(\eta\mathbf{I}+\mathbf{\Gamma}^{k}{\mathbf{\Gamma}^{k}}^{T})^{-1}{\mathbf{\Gamma}^{k}}{\mathbf{J}^{k}}^{T}\mathbf{J}^{k})
  \notag
\end{align}
We plug~\eqref{eq:sparsecode_svd} in~\eqref{eq::prop1_expand} and obtain:

\begin{align}
&  \|\mathbf{J}^{k} - \Delta\mathbf{D}^{k}\mathbf{\Gamma}^{k}\|_{F}^{2} - \|\mathbf{J}^{k}\|_{F}^{2} \notag \\
 = & tr(\mathbf{V}_{1}\mathbf{\Theta}\mathbf{V}_{1}^{T}{\mathbf{J}^{k}}^{T}\mathbf{J}^{k}\mathbf{V}_{1}\mathbf{\Theta}\mathbf{V}_{1}^{T}) - tr(2\mathbf{V}_{1}\mathbf{\Theta}\mathbf{V}_{1}^{T}{\mathbf{J}^{k}}^{T}\mathbf{J}^{k}) \notag \\
 = & - tr((2\mathbf{\Theta}-\mathbf{\Theta}^{2})\mathbf{V}_{1}^{T}{\mathbf{J}^{k}}^{T}\mathbf{J}^{k}\mathbf{V}_{1}) \notag \\
 = & - tr(\mathbf{Q}\mathbf{V}_{1}^{T}\mathbf{J}_{k}^{T}\mathbf{J}_{k}\mathbf{V}_{1}\mathbf{Q})  \\
 = &  - \|\mathbf{J}_{k}\mathbf{V}_{1}\mathbf{Q}\|_{F}^{2} \leq 0 \notag
\end{align}
where $\mathbf{Q} = diag(\mathbf{\overrightarrow{q}}) = diag(\frac{\sqrt{\lambda_{1}^{4}+2\eta\lambda_{1}^{2}}}{\lambda_{1}^{2}+\eta},...,\frac{\sqrt{\lambda_{n}^{4}+2\eta\lambda_{n}^{2}}}{\lambda_{n}^{2}+\eta})$ 

\ifCLASSOPTIONcaptionsoff
  \newpage
\fi

{\small
\bibliographystyle{ieee}
\bibliography{egbib_tip}
}

\begin{IEEEbiography}[{\includegraphics[width=1in,height=1.25in,clip,keepaspectratio]{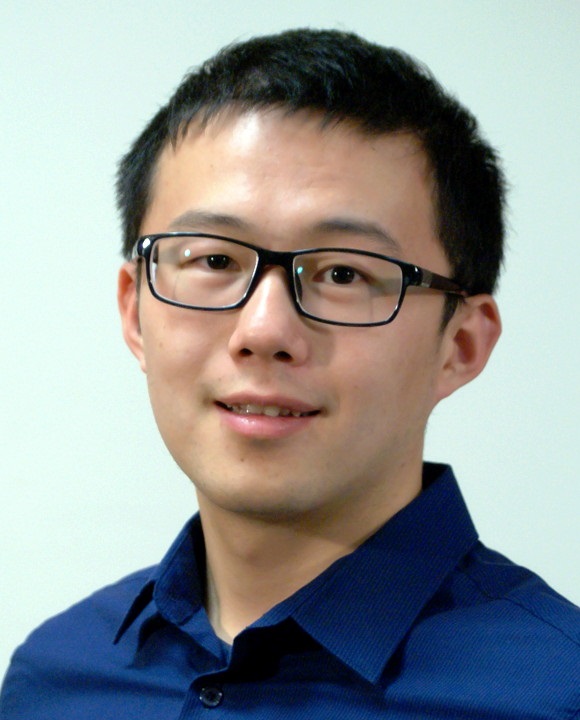}}]{Hongyu Xu}
received the B.E. degree from the University of Science and Technology of China in 2012 and the M.S. degree from University of Maryland, College Park in 2016. He is currently a research assistant in the Institute for Advanced Computer Studies at the University of Maryland, College Park, advised by Prof. Rama Chellappa. He is a former research intern with Snap Research (summer, fall 2017) and Palo Alto Research Center (PARC) (summer 2014). His research interests include dictionary learning, object detection, deep learning, face recognition, object classification, and domain adaptation.
\end{IEEEbiography}

\begin{IEEEbiography}[{\includegraphics[width=1in,height=1.25in,clip,keepaspectratio]{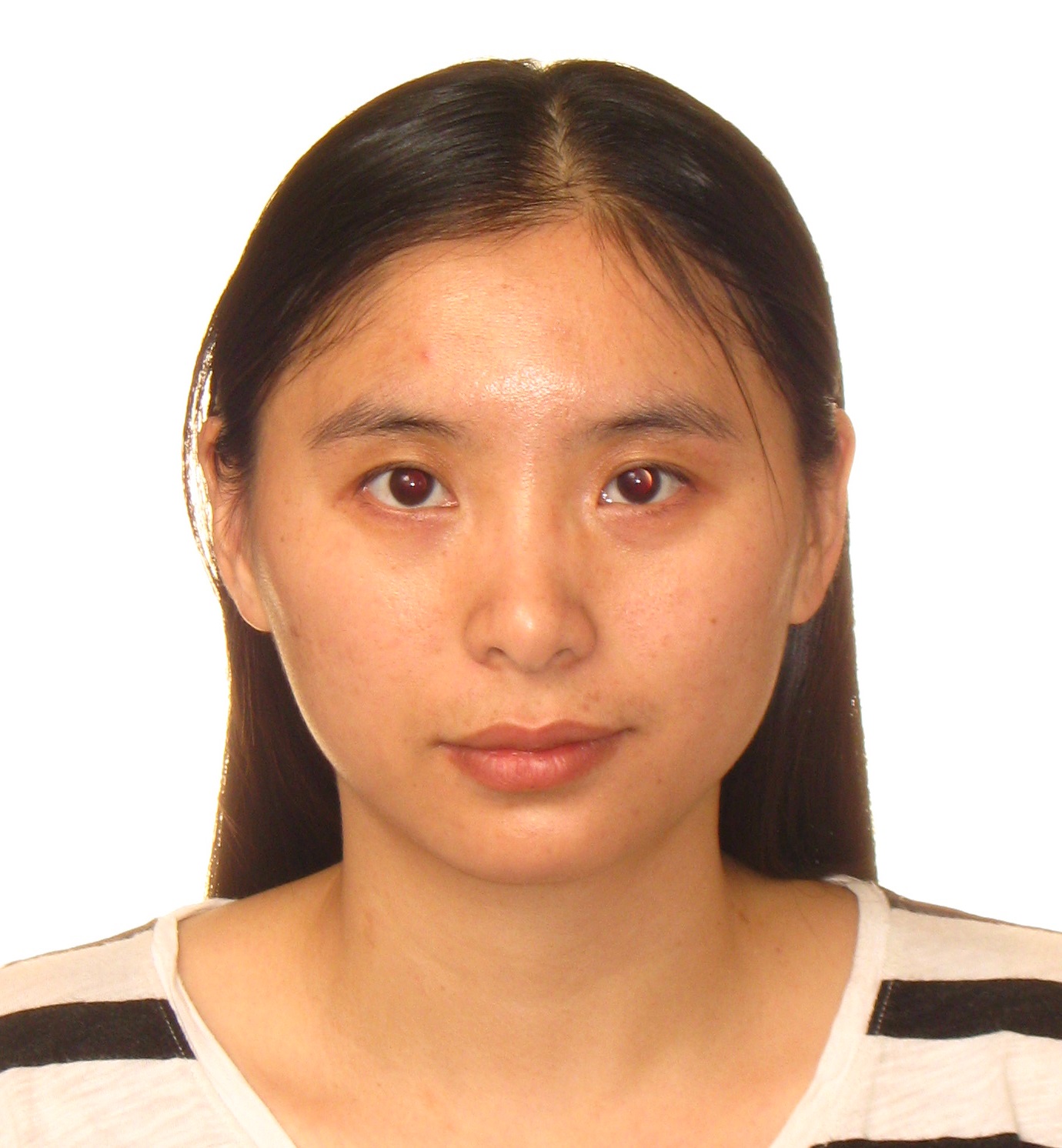}}]{Jingjing Zheng}
received the PhD degree in electrical and computer engineering from University of Maryland College Park in 2015. She is currently working as a computer vision scientist in General Electric Global Research. Her research interests include face identification, action recognition, and domain transfer learning.
\end{IEEEbiography}

\begin{IEEEbiography}[{\includegraphics[width=1in,height=1.25in,clip,keepaspectratio]{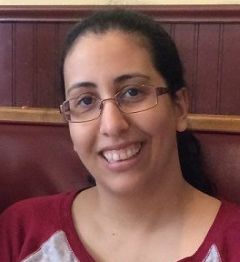}}]{Azadeh Alavi}
received the BS degree in applied Mathematics, and was awarded a PhD in computer vision and pattern recognition. She has served as a committee member of Australian Computer Society (Gold Coast chapter) for around 6 years, and has been working for more than 10 years as a researcher in both academia and industry (including 2 years and half as a research associate for University of Maryland). Her main interests are in image and video analysis, pattern recognition, machine learning and Neural Networks.
\end{IEEEbiography}


\begin{IEEEbiography}[{\includegraphics[width=1in,height=1.25in,clip,keepaspectratio]{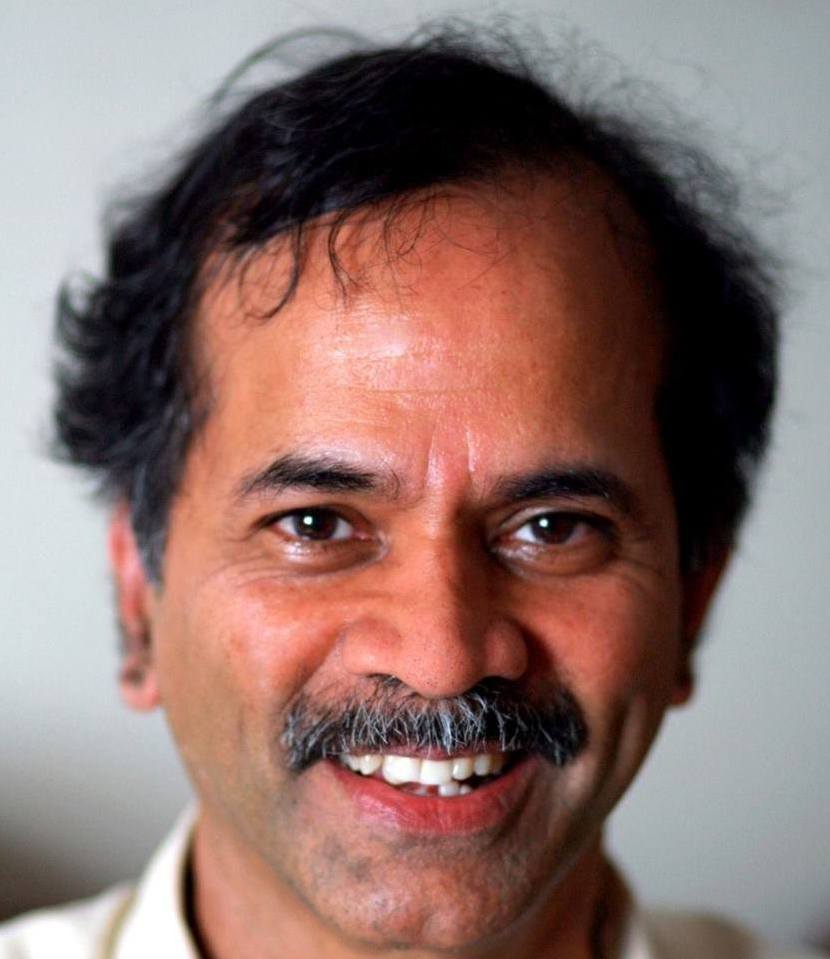}}]{Rama Chellappa}
received the BE (Hons.) degree in electronics and communication engineering from the University of Madras, India and the ME degree (with Distinction) in electrical communication engineering from the Indian Institute of Science, Bengaluru, India, in 1975 and 1977, respectively. He received the MSEE and PhD degrees in electrical engineering from Purdue University, West Lafayette, IN, in 1978 and 1981 respectively. During 1981-1991, he was a faculty member in the department of EESystems
at University of Southern California (USC). Since 1991, he has
been a professor of electrical and computer engineering (ECE) and an
affiliate professor of computer science at the University of Maryland
(UMD), College Park. He is also affiliated with the Center for Automation
Research and the Institute for Advanced Computer Studies (Permanent
Member) and is serving as the chair of the ECE Department. In 2005, he
was named a Minta Martin Professor of Engineering. His current
research interests span many areas in image processing, computer
vision and pattern recognition. He has received several awards including
a National Science Foundation (NSF) Presidential Young Investigator
Award and four IBM Faculty Development Awards. He received two
paper awards and the K.S. Fu Prize from the International Association of
Pattern Recognition (IAPR). He received the Society, Technical
Achievement and Meritorious Service Awards from the IEEE Signal
Processing Society. He also received the Technical Achievement and
Meritorious Service Awards from the IEEE Computer Society. He
received Excellence in teaching award from the School of Engineering
at USC. At UMD, he received college and university level recognitions
for research, teaching, innovation and mentoring of undergraduate students.
In 2010, he was recognized as an Outstanding ECE by Purdue
University. In 2016, he received the Outstanding Alumni Award from the Indian Institute of Science, Bangalore, India.
He served as the editor-in-chief of IEEE Transactions on Pattern
Analysis and Machine Intelligence and as the General and Technical
Program chair/co-chair for several IEEE international and national
conferences and workshops. He is a golden core member of the IEEE
Computer Society, served as a distinguished lecturer of the IEEE Signal
Processing Society and as the president of IEEE Biometrics Council.
He is a fellow of the IEEE, IAPR, OSA, AAAS, AAAI and ACM and holds four patents.
\end{IEEEbiography}


\end{document}